\documentclass{article} %
\usepackage{iclr2024_conference,times}

\usepackage{amsmath,amsfonts,bm}

\def\eqref#1{equation~\ref{#1}}
\def\Eqref#1{Equation~\ref{#1}}

\def\1{\bm{1}}

\DeclareMathAlphabet{\mathsfit}{\encodingdefault}{\sfdefault}{m}{sl}
\SetMathAlphabet{\mathsfit}{bold}{\encodingdefault}{\sfdefault}{bx}{n}

\usepackage{hyperref}
\usepackage{url}

\usepackage{enumitem}
\usepackage{xcolor} %
\definecolor{old_color}{RGB}{0,0,0}
\definecolor{new_color}{RGB}{0,0,0}
\definecolor{final_color}{RGB}{0,0,0}

\usepackage{graphicx}
\graphicspath{ {./images/} }

\usepackage{amssymb}

\title{The common Stability Mechanism behind most Self-Supervised Learning Approaches}

\author{%
  Abhishek Jha\thanks{\url{email: abhishek.jha@esat.kuleuven.be}}
  \\
  KU Leuven\\
  \And
  Matthew B. Blaschko \\
  KU Leuven\\
  \And
  Yuki M. Asano\\
  University of Amsterdam\\
  \And
  Tinne Tuytelaars\\
  KU Leuven\\
}

\iclrfinalcopy %
\begin{document}
\color{old_color}

\maketitle

\begin{abstract}
   Last couple of years have witnessed a tremendous progress in self-supervised learning (SSL), the success of which can be attributed to the introduction of useful inductive biases in the learning process to learn meaningful visual representations while avoiding collapse. These inductive biases and constraints manifest themselves in the form of different optimization formulations in the SSL techniques, e.g.~by utilizing negative examples in a contrastive formulation, or exponential moving average and predictor in BYOL and SimSiam. In this paper, we provide a framework to explain the stability mechanism of these different SSL techniques: i) we discuss the working mechanism of contrastive techniques like SimCLR, non-contrastive techniques like BYOL, SWAV, SimSiam, Barlow Twins, and DINO;  ii) we provide an argument that despite different formulations these methods implicitly optimize a similar objective function, i.e. minimizing the magnitude of the expected representation over all data samples, or the mean of the data distribution, while maximizing the magnitude of the expected representation of individual samples over different data augmentations; iii) we provide mathematical and empirical evidence to support our framework. We formulate different hypotheses and test them using the Imagenet100 dataset.
\end{abstract}

\section{Introduction}

Recent self-supervised learning (SSL) methods
aim for representations invariant to strong perturbations (called augmentations) of the input image. 
These perturbations are changes made to an input image that are supposed to preserve the underlying semantics.
Examples commonly used in SSL include random cropping, random rotation, color jittering, flipping and masking. These perturbations help the model learn to recognize the underlying structure of the image and its features, without being affected by irrelevant variations.
In practice, this is done by training a projector that maps different augmentations of the same image onto the same point in the feature space, and using the gradient of the loss (the distance between the representations) to train the projector.
These methods have shown to be highly effective in learning general features that can be transferred to a host of downstream tasks like classification \citep{van2020scan}, segmentation \citep{van2021unsupervised}, depth-estimation \citep{bachmann2022multimae}, and so on.

The objective of minimizing the distance between two augmentations of the same image can lead to a trivial solution, where all images are projected onto a single point in the feature space. This phenomenon is known as {\em embedding collapse} \citep{zhang2022does}. Different SSL techniques use different approaches to solve this problem: contrastive SSL techniques maximize the distance between an image and other images in the dataset (called negative pairs) while minimizing the distance between the image and its augmented versions (called positive pairs). They attribute the pulling force of positive pairs to learning invariance across different augmentations, and the pushing force of negative examples to collapse avoidance \citep{chen2020simple, he2019momentum}. 
\textcolor{new_color}{Non-contrastive methods do not require negative examples, and can be cluster-based, predictor-based, and redundancy minimization based.} Cluster-based non-contrastive SSL \citep{caron2020unsupervised} 
uses equipartitioning of cluster assignments for the collapse avoidance.
Another non-contrastive SSL method SimSiam \citep{chen2020exploring} uses an asymmetric student-teacher network with identical encoder architecture, and an additional projection layer, called predictor head, over the student to learn the SSL features. They claim the predictor learns the augmentation invariance. However, the exact collapse avoidance strategy of these methods is still unclear, with an empirical study pointing to a negative center vector gradient as a possible explanation \citep{zhang2022does}. Finally, redundancy-reduction based non-contrastive technique, Barlow Twins \citep{zbontar2021barlow}, uses redundancy-minimization through orthogonality constraints over the feature dimensions as a way to avoid collapse.

In this work, our goal is to uncover the underlying mechanisms behind these SSL methods. We show that they are actually instantiations of a common mathematical framework that balances training stability and augmentation invariance, as illustrated in Figure~\ref{fig:method}. We show this common framework motivates different hyperparameter and design choices that previously were set mostly empirically to obtain the best performance on downstream tasks. Our contributions are:

\begin{figure}
    \centering
    {\includegraphics[width=\textwidth]{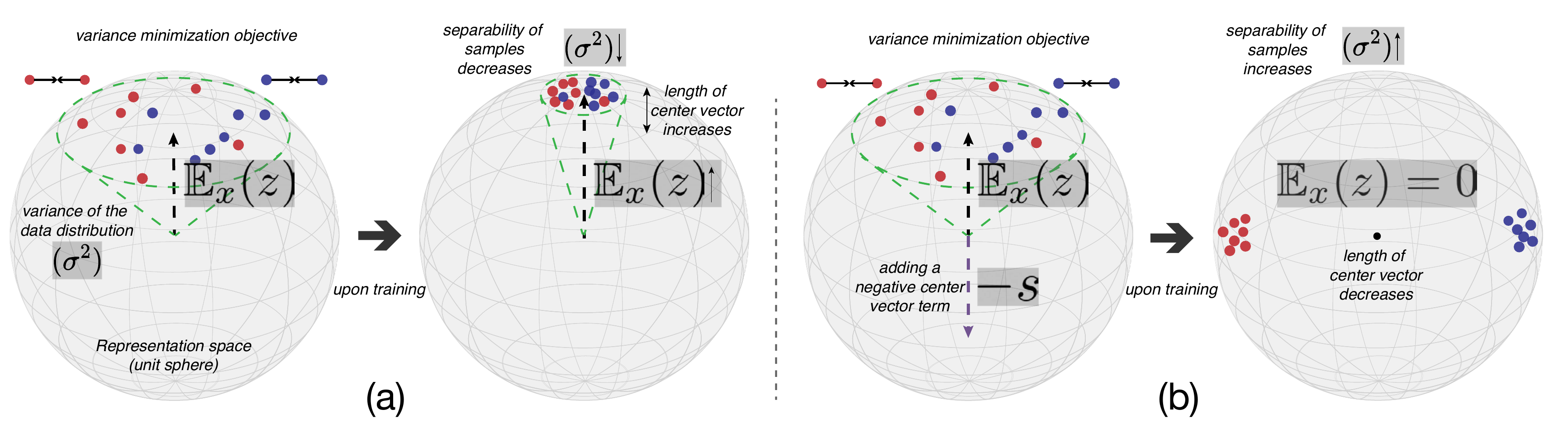}}
    \caption{\textbf{Overview of our proposed learning hypothesis.} Red and blue points represent different views of two images in feature space. (a) By applying distance minimization loss between two views of the same image, the magnitude of the expected representation over the data ($\mathbb{E}_x[z]$)  increases, reducing the variance of the data distribution ($\sigma^2$) in the feature space and thereby reducing their separability. (b) In order to learn a discriminative feature representation, a negative force ($-s$) equal to the expected representation over the data distribution is required. We hypothesize that this negative term is the collapse avoidance mechanism underlying different SSL methods. \vspace{-1.5em}}    \label{fig:method}
\end{figure}

\textcolor{new_color}{
Major contributions:
\begin{enumerate}[itemsep=0.25pt,leftmargin=*]
\item We propose a single framework/meta-algorithm that explains the underlying collapse avoidance mechanism behind contrastive and non-contrastive techniques. 
\begin{itemize}
\item Provide a simple mathematical formulation that can explain embedding collapse for distance minimization objective (also called invariance loss).
\item Reformulating all SSL techniques showing their mathematical conversion to our proposed framework, explaining that these techniques implicitly optimize our proposed framework to avoid embedding collapse.
\end{itemize}
\item We propose a simplistic technique based on our framework, that combines distance minimization with center vector magnitude minimization as a constraint optimization problem. This also provides an empirical justification for our proposed framework.
\end{enumerate}
Minor contributions:
\begin{enumerate}[itemsep=0.25pt,leftmargin=*]
\item We explain peculiar cases of existing SwAV with fixed prototype and Barlow twins without off-diagonal minimization in the purview of our framework.
 \item We show that our proposed algorithm can be used to make predictions about, and understand rationality behind, some hyper-parameter selection in these SSL techniques which are otherwise selected purely empirically.
\end{enumerate}
}

\textit{Scope:} We define the scope of our work under which we explore self-supervised learning:
\begin{enumerate}[itemsep=0.25pt,leftmargin=*]
\item We explore contrastive and non-contrastive methods of learning representation, and do not discuss Masked-image-models \citep{bao2021beit} or other methods based on proxy task such as rotation \citep{gidaris2018unsupervised}, colorization \citep{zhang2016colorful}, jigsaw \citep{noroozi2016unsupervised}, relative patch location \citep{doersch2015unsupervised} etc., as they are not optimizing the feature space directly and unlikely to suffer from collapse as the ones we discuss in this paper.
\item The objective to learn the representation being invariance loss: We do not consider equivariance objectives as contemporary methods in self-supervised learning use invariance loss to minimize distance between augmented version of the input.
\end{enumerate}

\section{Formulation}
Despite the differences in the design choices and approaches, there is a unifying principle behind different SSL methods. This principle can be divided into two objectives: first, learning the augmentation invariance of the images, and second, ensuring stability in the representation space by avoiding embedding collapse. We propose that the key to stability lies in constraints imposed on the expected representation over the dataset, or what we call the {\em center vector}, a term coequally used by \citet{zhang2022does}. These constraints prevent SSL methods from converging to trivial solutions. In particular, architectures where the optimization function minimizes the magnitude of the center vector avoid collapse, while the ones where the optimization function does not constrain it, collapse to trivial solutions. 
In this section, we provide a mathematical framework based on augmentation invariance and center vector constraints that generalizes different SSL approaches. We later redefine these different approaches in the purview of our framework.

\begin{figure}[t]
  \begin{center}
      \includegraphics[width=\linewidth]{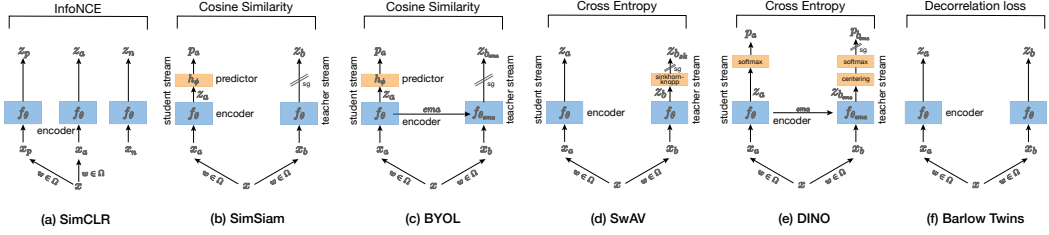}
   \end{center}
      \caption{All methods covered by our proposed framework. For details please zoom-in.}
      \label{fig:all_methods}
   \end{figure}

\subsection{Two-stream Self-supervised Learning: The Role of the Center Vector}
\label{sec:cv_section}
\textcolor{new_color}{Let an encoder function $f$ map the RGB image space $x \in \mathcal{I}$ to the representation space $\mathbb{R}^D$ which is then normalized to the unit sphere $z \in \mathcal{S}^D$, $z = g(x) = f(x)/\|f(x)\|_2$, and $\omega : \mathcal{I} \rightarrow \mathcal{I}$ be a stochastic augmentation with distribution $P_\omega$. The center vector, $s$, is defined as the expected representation over the augmented input distribution:}
\begin{align}
s := \mathbb{E}_{x\sim P_x}\left[\mathbb{E}_{\omega\sim  P_\omega}\left[g(\omega(x))\right]\right]
\label{eq:define_center_vector}
\end{align}

For different augmentations, $\omega \in \Omega$, a two-stream self-supervised objective minimizes the distance, $\mathcal{D}$, between $z$ and the representation of the augmented version of $x$:
\begin{align}
L(f) = \mathbb{E}_{x,\omega} \left[\mathcal{D}(z,z_{\omega}) \right]
= \mathbb{E}_{x,\omega} \left[ \mathcal{D}(g(x),g(\omega(x))) \right] .
\label{eq:define_objective_base}
\end{align}
where $z_\omega$ denotes the representation of the augmented view of the data. Here, we define our framework:

\textbf{Framework:}
Optimizing only the augmentation invariance objective defined in equation \ref{eq:define_objective_base}, may lead to a trivial solution, as the representations, $z$, collapse. 
We posit that a non-trivial solution can be obtained by selecting an objective function that constrains the expected global representation to zero:
\begin{align}
{\min}\; L(f);\;\; s.t. \;\;  s = 0
\label{eq:define_objective_hat_s}
\end{align}

\textbf{Explanation:} The main objective function of distance minimization, $\mathcal{D}$, between the two views of the data is usually the Euclidean distance squared $\|z-z_{\omega}\|_2^2$, which is equivalent to minimizing the negative cosine similarity,$-\langle z, z_{\omega}\rangle$, for $L_2$ normalized vectors. Hence, the loss gradient with respect to the feature vector $z$ can be written as $\frac{\partial}{\partial z}L(f) = -\mathbb{E}_\omega [z_{\omega}]$. As the representation vector $z$ will move in the opposite direction of the gradient to minimize the loss, $z$ will move in the direction of $\mathbb{E}_\omega [z_{\omega}]$.  Note that we have not considered yet the constraint $s=0$ as $s$ must be estimated from a (stochastic) finite sample and the constrained optimization is difficult in general. %

Let the expected value of $z$ over augmentations $\omega\in \Omega$, is $\mu_z = \mathbb{E}_{\omega}[z_{\omega}]$. Therefore, each term in the sum of equation \ref{eq:define_objective_base}, 
can be rewritten 
as:

\begin{align}
\frac{\partial}{\partial z}L(f) = -\mathbb{E}_{x}[\mu_z] - (\mathbb{E}_{\omega}[z_{\omega}] - \mathbb{E}_{x}[\mu_z])
\label{eq:delLx_Ez_z_Ez}
\end{align}

The first term in the above equation leads $z$ to move in the direction of the batch center, which is common across all the samples in the batch. The second term leads it to the residual direction which is different for different samples. As $z_{\omega}$ is $L_2$ normalized, the magnitude of the sum of these terms is bounded above by $1$, 
so in expectation the larger the magnitude of the expected representation, $\mathbb{E}_{x}[\mu_z]$ (first term) is, the smaller the magnitude of the residual representation, $\mathbb{E}_{\omega}[z_{\omega}] - \mathbb{E}_{x}[\mu_z]$ (second term) will be, which is not desirable. As $z$ keeps moving in the direction of the expected batch representation, its value iteratively increases and the residual vectors become smaller and smaller. 
All this can be avoided by minimizing the magnitude of the center vector, or the expected batch representation, $\mathbb{E}_{x}[\mu_z] = 0$, resulting in larger residual terms, $(\mathbb{E}_\omega [z_{\omega}] - \mathbb{E}_{x}[\mu_z])$. This residual term is different for different samples, and hence represents the semantic content of the sample. For a large batch size, the samples are representative of the global distribution of the dataset, hence the batch center $\hat{s}$  coincides with the dataset center, $\mathbb{E}_{x}[\mu_z]$. We define the residual component for a sample $z$ as $r_{z} = (z - \mathbb{E}_{x}[\mu_z])$.

For each iteration, explicit computation of the global center vector is non-trivial and expensive. Instead, different SSL approaches employ different ways to  incorporate center vector minimization, despite being non-explicit about it. Using our framework of center vector minimization, we redefine the contemporary SSL approaches.

\subsection{Contrastive SSL: Role of negative examples}\label{sec:ConstrastiveSSL}
\textbf{Triplet loss:}
In contrastive SSL approaches, the goal is to bring the representations of the augmented views of an input sample close while pushing away that of other samples.
To study constrastive SSL, we formulate a triplet objective function \citep{hoffer2015deep}. For a standard triplet loss setup, as shown in Figure~\ref{fig:all_methods}a, $x_a$, an anchor, and $x_p = \omega(x_a)$, a positive exemplar, constitute the two views of the same data point $x_a$ and are called a positive pair, while $x_n$, a negative exemplar, is another data point and together with $x_a$ constitutes a negative pair. $z_a, z_p,$ and $z_n$ are their projections in the representation space, respectively.
Then triplet loss can be written as:
\begin{align}
L_{\text{triplet}}(f) = \mathbb{E}_{x_n,x_a,\omega} \left[ \frac{1}{2}{\max\big(\|z_a - z_p\|_2^2 - \|z_a-z_n\|_2^2 + \alpha, 0 \big)} \right]
\label{eq:L_triplet}
\end{align}

In self-supervised learning, typically the two different images are pushed as far apart as possible, hence the margin $\alpha \rightarrow \inf$, which is equivalent to
$%
\mathbb{E}_{x_n,x_a,\omega} \left[  \frac{1}{2}\big(\|z_a - z_p\|_2^2 - \|z_a-z_n\|_2^2 \big) \right]$ and
\begin{align}
L_{\text{triplet}}(f) =&\mathbb{E}_{x_n,x_a,\omega} \left[  -\langle z_a, z_p \rangle +\langle z_a, z_n \rangle  \right],
\label{eq:L_za}
\end{align}
where the equivalence is due to the $L_2$ normalization and constants have been removed from the objective.
To understand how the representation $z$ evolves, we analyze how anchors  move in the representation space. To check this, we can look at the gradient of the loss w.r.t. the anchor, $z_a$. The anchor then moves in the opposite direction of this gradient.

\begin{align}
\frac{\partial}{\partial z_a}L_{\text{triplet}}(f) = \mathbb{E}_{x_n,\omega} \left[-z_p + z_n\right]
= \mathbb{E}_{x_n,\omega}\left[\mathbb{E}_{x}[\mu_z] -(z_p - \mathbb{E}_{x}[\mu_z])  - \mathbb{E}_{x}[\mu_z] + (z_n - \mathbb{E}_{x}[\mu_z]) \right]\\
= \mathbb{E}_{x_n,\omega}\left[-s -r_p + s + r_n \right] = \mathbb{E}_{x_n,\omega}\left[-r_p  + r_n \right]
\label{eq:partial_za_L_2}
\end{align}
$-r_p + r_n$ is the difference of the residual vectors for the positive and negative exemplars, respectively, and is desirable as it would move $z_a$ in the direction of $r_p$, the semantic component of the representation of the positive sample, and away from that of the negative sample $r_n$. 
If we did not have a negative sample term, $z_n$, the loss gradient would be exactly what we had in equation \ref{eq:delLx_Ez_z_Ez}. Eventually, the center vector $s$ would become very large compared to $r_p$, as an increase in the center vector leads to a decrease in the residual as their sum is upper bounded by 1. In this case, all samples in the dataset will have high similarity to each other, since the $s$ component is present in all of them, and the difference between any two samples, $x_i$ and $x_j$ would be small. This eventually causes collapse.

\textbf{InfoNCE:}
In practice, most contrastive SSL methods use the InfoNCE Loss:
\begin{align}
L_{\text{InfoNCE}}(f) =& -\mathbb{E}_{x_n,x_a,\omega} \left[\log\left(\frac{\exp(\text{sim}(z_a,z_p)/\tau)}{\exp(\text{sim}(z_a,z_p)/\tau) + \sum_{z_n}\exp(\text{sim}(z_a,z_n)/\tau)}\right)  \right]\\
=& -\mathbb{E}_{x_n,x_a,\omega} \left[\text{sim}(z_a,z_p)/\tau - \operatorname{LogSumExp}(\text{sim}(z_a,z_p)/\tau,
 \{\text{sim}(z_a,z_n)/\tau\}) \right] .
\label{eq:cont_ssl_L_Linfo}
\end{align}

In SimCLR \citep{chen2020simple}, a very low temperature
$(\tau \ll 1)$ is used. Using the identity, $\lim_{\tau \searrow 0} \tau \operatorname{LogSumExp}(\{\text{sim}(z_a,z_i)/\tau\}) = \max_i \text{sim}(z_a,z_i)$, we see that the objective approaches %
\begin{align}
L_{\text{InfoNCE}}(f) \approx -\mathbb{E}_{x_n,x_a,\omega} \left[\log\left(\frac{\exp(\text{sim}(z_a,z_p)/\tau)}{ \exp(\text{sim}(z_a,z_{n_{max}})/\tau)}\right)\right]\\
= -\mathbb{E}_{x_n,x_a,\omega} \left[\text{sim}(z_a,z_p)/\tau - \text{sim}(z_a,z_{n_{max}})/\tau \right].
\label{eq:cont_ssl_L_neg_max_simple}
\end{align}
The similarity function in equation \ref{eq:cont_ssl_L_neg_max_simple} most commonly used in the literature has been cosine similarity, and ignoring the constant $\tau$, the resulting objective is equivalent to:
\begin{align}
L_{\text{InfoNCE}}(f) \approx -\mathbb{E}_{x_n,x_a,\omega} \left[\langle z_a,z_p \rangle - \langle z_a,z_{n_{max}} \rangle \right].
\label{eq:cont_ssl_L_neg_max_cosine}
\end{align}

In summary, for normalized representation vectors, equation \ref{eq:cont_ssl_L_neg_max_cosine} becomes equivalent to equation \ref{eq:L_za}. Hence, for InfoNCE loss as well, the stability of the representation depends on the constraint over the magnitude of the center vector.
Equation \ref{eq:partial_za_L_2}, and the paragraph following it provide the role of positive examples for feature invariance maximization and negative examples for collapse avoidance. 
Further, equation \ref{eq:cont_ssl_L_neg_max_simple} shows the use of temperature as a measure to sample hard-negatives. %

\subsection{SimSiam: How predictor helps avoid embedding collapse}
\label{sec:simsiam}
Given the SimSiam setup in Figure~\ref{fig:all_methods}b: %
$x_a = \omega_a(x)$ and $x_b = \omega_b(x)$ are two augmentations of x subjected to augmentations $\omega_a \sim P_\omega$ and $\omega_b \sim P_\omega$, respectively. $f_\theta : \mathcal{X} \rightarrow \mathcal{S}^D$ is an encoder function, parameterized by $\theta$. $h_\phi : \mathcal{S}^D \rightarrow \mathcal{S}^D$ is a predictor function, parameterized by $\phi$, such that $z_a = f_\theta(x_a)$, $z_b = f_\theta(x_b)$ and $p_a = h_\phi(z_a) = h_\phi(f_\theta(x_a))$. For any iteration $t$, these three equations and the corresponding loss are:
\begin{align}
z_a^t = f_\theta^t(x_a);\;\;
z_b^t = f_\theta^t(x_b);\;\;
p_a^t = h_\phi^t(z_a) = h_\phi^t(f_\theta^t(x_a))
\label{eq:simsiam_z_a_t}\\
L_{\text{SimSiam}}^t(f,h) = \frac{1}{2} \mathbb{E}_{x,\omega_a,\omega_b} \left[-\langle p_a^t, \texttt{sg}(z_b^t) \rangle - \langle p_b^t , \texttt{sg}(z_a^t) \rangle \right] 
\label{eq:simsiam_L_t} \\
= \mathbb{E}_{x,\omega_a,\omega_b} \left[-\langle p_a^t, \texttt{sg}(z_b^t) \rangle \right]
\end{align}
where $\texttt{sg}$ indicates that backpropagation will not proceed on that variable \citep{chen2020exploring}.  It holds
\begin{align}
2-2\langle p_a^t, \texttt{sg}(z_b^t) \rangle = \|p_a^t - \texttt{sg}(z_b^t)
\|_2^2 = \|h_\phi^t(z_a^t)- \texttt{sg}(z_b^t)\|_2^2
\label{eq:simsiam_L_t_h}
\end{align}
Now let us have a look at the predictor alone, as shown in Figure~\ref{fig:all_methods}. Since $h_\phi^t$ has been optimized in the $t-1$ backward pass, it minimized the loss term $L_{ab}^{t-1} := \|p_a^{t-1} - \texttt{sg}(z_b^{t-1})
\|_2^2 $.
The update of $h_\phi$ at the $t-1$ iteration is $\phi^t = \phi^{t-1} -\lambda\frac{\partial}{\partial\phi_{t-1}}L_{ab}^{t-1}$.
SimSiam uses a high learning rate ($\lambda$) for the predictor to update it more frequently. Hence, $h_\phi$ learns to project $z_a$ to $z_b$ almost perfectly, and after the update of $h_\phi^{t-1}$ to $h_\phi^t$, we have %
\begin{align}
h_\phi^t(z_a^{t-1}) \approx z_b^{t-1}
\label{eq:simsiam_p_a_t-1}
\end{align}

After the $t-1$st update of $f_\theta$, the new updated encoder $f_\theta^t$ projects $x_a$ and $x_b$ to $z_a^t$ and $z_b^t$ respectively. This causes a shift of distribution from $P(z|\theta^{t-1},x)$ to $P(z|\theta^t,x)$,
due to change in parameters from $\theta^{t-1}$ to $\theta^t$, which we denote $\mathbb{E}_{x,\omega_a}[f_\theta^t(x_a) -f_\theta^{t-1}(x_a)] = \mathbb{E}_{x,\omega_a}[\mathcal{\vec{D}}(z_a^{t-1},z_a^t)] =: \Delta_{\text{dist}}^t$. %

\begin{align}
\because f_\theta^t(x_a) = z_a^t = z_a^{t-1} + \vec{\mathcal{D}}(z_a^{t-1},z_a^t)
\label{eq:simsiam_f_t_x_a}\\
\Rightarrow L_{ab}^t= \|h_\phi^t(z_a^{t-1} + \vec{\mathcal{D}}(z_a^{t-1},z_a^t)) - \texttt{sg}(z_b^t)\|_2^2
\label{eq:simsiam_L_t_h_phi}
\end{align}

Since the predictor is trained to adapt quickly to the encoder, with high learning rate \citep{chen2020exploring}, we assume that $h_\phi^t$ is invariant to small changes $z$:
\begin{align}
\therefore h_\phi^t(z_a^{t-1}+\mathcal{\vec{D}}(z_a^{t-1},z_a^t))  \approx
     h_\phi^t(z_a^{t-1})
\label{eq:simsiam_h_phi_t_z_a}\\
\therefore \text{expected loss becomes}:\;\;\ \mathbb{E}_{x,\omega_a,\omega_b}[L_{ab}^t] \approx \mathbb{E}_{x,\omega_a,\omega_b}[\|h_\phi^t(z_a^{t-1})-\texttt{sg}(z_b^t)\|_2^2] \\
\approx \mathbb{E}_{x,\omega_b}[\|z_b^{t-1}-\texttt{sg}(z_b^t)\|_2^2]
\label{eq:simsiam_z_b_t-1}\\
\therefore \text{where the last approximation is from eq. }\ref{eq:simsiam_p_a_t-1}:\;\; L_{\text{SimSiam}}^t(f,h)=\mathbb{E}_{x,\omega_a,\omega_b}[L_{ab}^t]\approx \|\Delta_{\text{dist}}^t\|_2^2
\label{eq:simsiam_L_t_dist}
\end{align}

Change in the distribution between $z^{t-1}$ to $z^t$ can be written as the shift in their mean 
\begin{align}
\|\Delta_{\text{dist}}\|_2^2 \propto %
\|\mathbb{E}_{x,\omega}[z_\omega^t] - \mathbb{E}_{x,\omega}[z_\omega^{t-1}]\|_2^2
\label{eq:simsiam_Delta_dist}
\end{align}

Equation \ref{eq:simsiam_Delta_dist} suggests the expected loss at iteration $t$, causally depends on the expected representation in iteration $t-1$. We can extrapolate this to the $t=0$, where for a randomly initialized representation space, $\mathbb{E}_{x,\omega}[z_\omega^0] \approx 0$. This causal reliance on previous iteration acts as a constraint in limiting the increase of center vector in iteration $t$.

\subsection{BYOL: Role of exponential moving average}
\label{sec:BYOL}
Similar to SimSiam, BYOL \citep{grill2020bootstrap} is also a two-stream network with predictor on top of the online stream, Figure~\ref{fig:all_methods}c. However, unlike SimSiam, the offline network is updated as the exponential moving average (EMA) of the online stream. Secondly, while SimSiam requires the expedite learning of the predictor with a high learning rate, BYOL uses a smaller learning rate of predictor. As a high learning rate is critical for such an architecture to avoid collapse, while BYOL still manages to avoid it, the EMA should be playing an important role in collapse avoidance.

Let the online network be $f_\theta$ and the offline network be $f_{\theta_{\text{ema}}}$, with parameter $\theta_{\text{ema}}$ updated as $\theta_{\text{ema}}^t = (1-\epsilon) \theta^t + \epsilon \theta_{\text{ema}}^{t-1}$, with a typical value of $\epsilon = 0.99$ \citep{grill2020bootstrap}. The resulting outputs of the two views of the data points, $x_a$ and $x_b$, are $z_a$ and $z_{b_{\text{ema}}}$. Here, we can rewrite
$z_{b_{ema}} = \epsilon f_{\theta_{ema}}^{t-1}(x_b)
 + z_{b_{ema}} - \epsilon f_{\theta_{ema}}^{t-1}(x_b) =: \epsilon f_{\theta_{ema}}^{t-1}(x_b) + z_{b_\delta}^t$. 
\begin{align}
L_{\text{BYOL}}(f_\theta,f_{\theta_{ema}}) = -\mathbb{E}_{x,\omega_a,\omega_ b}\left[ \langle z_a, \texttt{sg}(z_{b_{\text{ema}}})\rangle \right];\;\; 
\frac{\partial}{\partial z_a}L_{\text{BYOL}} = - \mathbb{E}_{\omega_b}\left[
\epsilon %
z_{b_{ema}}^{t-1}
+ z_{b_\delta}^t \right] \\
=
-\mathbb{E}_{\omega_b}\left[\epsilon \left(r_{z_{b_{ema}}}^{t-1} + \mathbb{E}_{x}[\mu_{z_{b_{ema}}}^{t-1}]\right)
+  r_{z_{b_{\delta}}}^t + \mathbb{E}_{x}\left[\mu_{z_{b_{\delta}}}^t\right]
\right] \\
\approx -\mathbb{E}_{\omega_b}\left[\epsilon \left(r_{z_{b_{ema}}}^{t-1} + \mathbb{E}_{x}[\mu_{z_{b_{ema}}}^{t-1}]\right)
+  r_{z_{b_{\delta}}}^t \right]
\end{align}
We see that $z_{b_\delta}^t = z_{b_{ema}} - \epsilon f_{\theta_{ema}}^{t-1}(x_b)$ will in general be close to zero, as $f_{\theta_{ema}}^{t-1}$ is close to $f_{\theta_{ema}}^{t}$ and $\epsilon$ is close to 1.
If the initial distribution of the features are randomly distributed across the unit hypersphere, the magnitude of the center vector is $0$, which the feature of each sample tries to move closer to in each subsequent iteration. This also means that no negative center vector is required in terms of the batch-level normalization etc, although they can improve the overall performance \citep{richemond2020byol}.
This momentum component only partially helps in minimizing the magnitude of the center vector, as discussed in the previous section, in SimSiam, the predictor helps in minimizing the magnitude of center vector as well, following a similar path of confining to the initial distribution of the feature space, that is centered to origin. It has also been shown in the literature \citep{richemond2020byol} that when the network is non-uniformly initialized leading to the non-uniform feature space, batch normalization becomes necessary to mitigate collapse. This can be explained using our derivation, that when the initial feature space is non-uniform, the center vector magnitude is non-zero, hence a negative center vector term is required to nullify the effect of $\mathbb{E}_{x}\left[\mu_{z_{b_{ema}}}^{t-1}\right]$.

\subsection{DINO}
\label{sec:DINO}
DINO, a two-stream network, uses a student-teacher network similar to BYOL, with teacher parameters updated as the exponential moving average of the student, Figure~\ref{fig:all_methods}e.
Architectures of both streams are symmetric without any predictor on student network. 
For the output representations $z_a, z_{b_{\text{ema}}}$ corresponding to student and teacher network respectively, we can write the loss as
\begin{align}
L_{\text{DINO}}(f) = -\mathbb{E}_{x,\omega_a,\omega_b}\left[\langle \texttt{sg}(\tau\text{softmax}((z_{b_{\text{ema}}}-C)/\tau)), \tau\log(\text{softmax}(z_{a}/\tau))\rangle \right],
\end{align}
where $C$ is a momentum term based on the expected value of $z$ \citep[Equation~(4)]{caron2021emerging}.  As above, taking the limit as $\tau$ goes to zero from above:
\begin{align}
    \lim_{\tau \searrow 0} L_{\text{DINO}}(f) = -\mathbb{E}_{x,\omega_a,\omega_b}\left[\langle \texttt{sg}(e_{\max(z_{b_{\text{ema}}}-C)}), z_a - \max(z_{a})\rangle \right] \\
    \frac{\partial}{\partial z_a} \lim_{\tau \searrow 0} L_{\text{DINO}}(f) = -\mathbb{E}_{\omega_b}\left[ e_{\max(z_{b_{\text{ema}}}-C)} \right] + e_{\max z_a} 
\end{align}
since $C$ is the exponential moving average of the batch centers over different iterations, this captures the notion of center vector the  $s_{b_{\text{ema}}}$. Hence the above equation becomes $ \frac{\partial}{\partial z_a}L_{\text{DINO}} \approx r_{b_{\text{ema}}} $. This gradient equation has a low center vector magnitude and hence the features do not move towards any certain direction and collapse is avoided. The importance of centering operation for collapse avoidance has also been studied in DINO.
With this reformulation, we reexamine the centering operation in the purview of our framework to explain why it helps in collapse avoidance.

\textbf{Note:} We find that the stability of SwAV \citep{caron2020unsupervised} and Barlow Twins \citep{zbontar2021barlow}, can also be explained through the center vector framework. Due to space constraint, we move the corresponding formulation sections to supplementary.

\section{Experiments}
\subsection{Simplified SSL objective: Penalizing center vector magnitude}
\begin{figure}[h]
    \centering
    {\includegraphics[width=\textwidth]{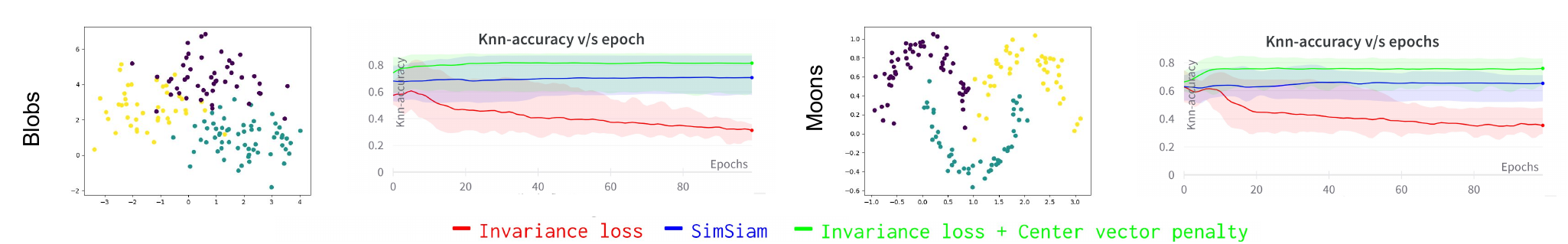}}
    \caption{\textbf{Simplified SSL objective:} We show that a simplified objective that minimizes the invariance loss with a center vector penalty (green), can outperform SimSiam. We plot the toy dataset distribution on left and performance curves on right for Blob and Moons dataset. Plots are averages of five runs with varying seeds, and variance is shown by shaded regions.
    }
    \label{fig:cv_penalty_Vs_simsiam}
\end{figure}

Based on the constrained optimization problem proposed in \Eqref{eq:define_objective_hat_s} we propose a simplified SSL objective:
\begin{align}
    L_{Simple}(f) =  0.5(L(f) - \lambda_\mathbb{L} s)
\end{align}
where $\lambda_L$ is the Lagrange multiplier, and act as a penalty term for minimizing the center vector. We optimize this unconstrained objective, $\text{min} L_{Simple}(f)$, through mini-batch SGD. In Figure~\ref{fig:cv_penalty_Vs_simsiam}, we compare the performance of this simplified objective against SimSiam on toy datasets. $\lambda_L$ is a hyperparameter which we set to $-1$, however a better selection process should be possible, but is beyond the scope of this paper. We observe that our simplified objective without architectural complexity of SimSiam, is able to outperform its performance. This provides a possible justification of the proposed framework in Section~\ref{sec:cv_section}.

\subsection{Why SimSiam collapses without predictor? Understanding collapse with toy-datasets}
\label{sec:simsiam_toy}
\begin{figure}[h]
  \begin{center}
      \includegraphics[width=\linewidth]{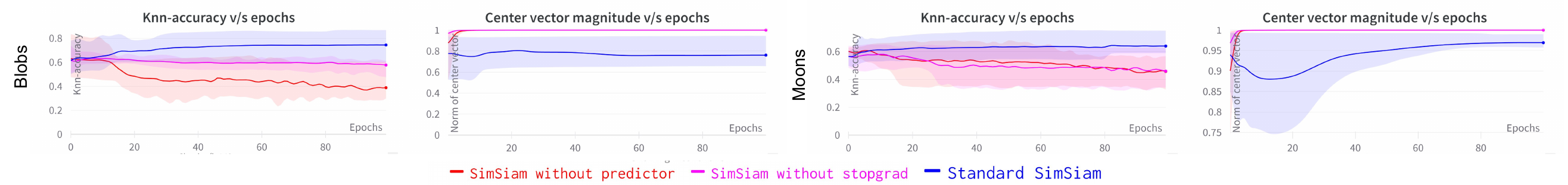}
   \end{center}
      \caption{\textbf{Evaluation on toy datasets:} Standard SimSiam,
      SimSiam without predictor and SimSiam without stop-gradient have been shown in blue, red and pink respectively. Plots are averages of five runs with varying seeds, and variance is shown by shaded regions. Center vector is high for both the cases of collapse, i.e. SimSiam without predictor, and SimSiam without stop-gradient. This empirically verifies, the role of predictor and stop-gradient for collapse avoidance in SimSiam, based on our formulation. Input dataset distribution can be viewed in Figure~\ref{fig:cv_penalty_Vs_simsiam}}
      \label{fig:toy_datasets}
   \end{figure}

Toy datasets provide a controlled abstraction over the complexity of the natural distribution and hence act as a test-bench for the empirical evaluation of our proposed framework. We incorporate two toy datasets: blobs, moons. Each of the datasets contains samples in 2 dimensional space for three classes, as shown in Figure~\ref{fig:toy_datasets}. We treat samples of one class as the augmentations of a single image and train a SimSiam model with and without 
stop-gradient.
Simsiam without predictor, and stop-gradient collapse for natural distribution \citep{chen2020exploring}
and acts as a good model to showcase the behavior of center vector for sub-optimal cases. In both SimSiam without predictor, and without stop-gradient cases,  the formulation defined in equation \ref{eq:simsiam_p_a_t-1}, and \ref{eq:simsiam_Delta_dist} do not hold. Hence, no center vector minimization term is present in the loss, leading to collapse. Analysis on natural dataset has been provided in supplementary.

\subsection{Barlow-twins can work without orthogonalization}
\label{sec:byol_ortho}

\begin{figure}[h]
    \centering
    {\includegraphics[width=0.8\textwidth]{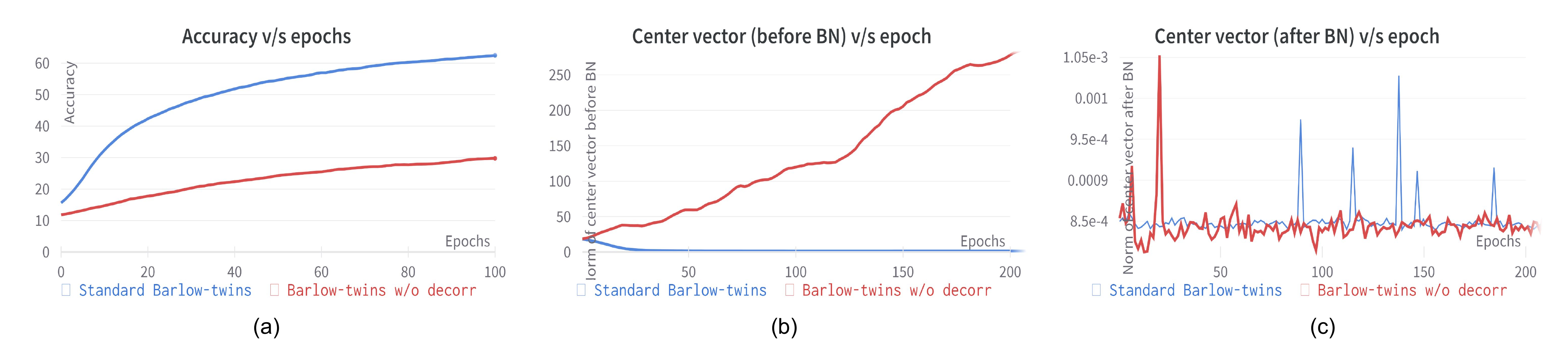}}
    \caption{Barlow-twins can learn non-collapse features without decorrelation term in the loss formulation: (a) shows the knn accuracy of Barlow-twins with and without the decorrelation terms in the loss on Imagenet100, (b) and (c) show the norm of the center vector of $z$ before and after BN, while training Barlow-twins in the two aforementioned settings, respectively. We can see that BN helps in removing the center vector component from $z$.
    }
    \label{fig:BT_no_ortho}
\end{figure}

Barlow-twins orthogonalizes cross-correlation matrices between image views. It uses an invariant loss for diagonal elements and an orthogonalization/decorrelation loss for off-diagonal elements, multiplied by a weighting factor $\lambda$. \citet{zbontar2021barlow} demonstrate the robustness of Barlow-twins to different $\lambda$ values. In our formulation in supplementary Section~\ref{sec:barlow_twins}, we show that pushing the off-diagonal elements to zero, is the same as minimizing the negative pair similiarity in the mini-batch. Hence, $\lambda$ should play a similar (however weaker role) as $\tau$, the temperature parameter in InfoNCE. While in InfoNCE, the $\tau$ parameter helps in sampling hard-negatives, there is no such mechanism to do so here with $\lambda$, and as per our section on Contrastive learning, hard-negative sampling is critical in minimizing the center vector magnitude and therefore in avoiding collapse. Hence, there must be some mechanism to compensate for the lack of hard-negative sampling in Barlow-twins to minimize the center vector. We argue, that Batch-normalization (BN) coupled with large batch-size in Barlow twins, helps in estimating the dataset center vector and its removal through BN. We also argue this is the reason behind the robustness against the $\lambda$ parameter in the original Barlow-twins formulation \citep{zbontar2021barlow}. To verify our claim, we train a Barlow-twins network without decorrelation/orthogonalization term, i.e. we only train the invariance term between two views of the data. Similar to the original implementation, we use BN. As shown in Figure~\ref{fig:BT_no_ortho}, we see that even without decorrelation-term, or in terms of InfoNCE equivalent, without any negative pairs, Barlow-twins is able to avoid collapse, due to BN, although with suboptimal performance. This additionally empirically verifies our proposed framework of center vector minimization for SSL stability.

\subsection{SwAV with fixed prototypes}
\begin{figure}[ht]
\centering
\begin{minipage}[b]{0.5\linewidth}
\centering
\includegraphics[width=\linewidth]{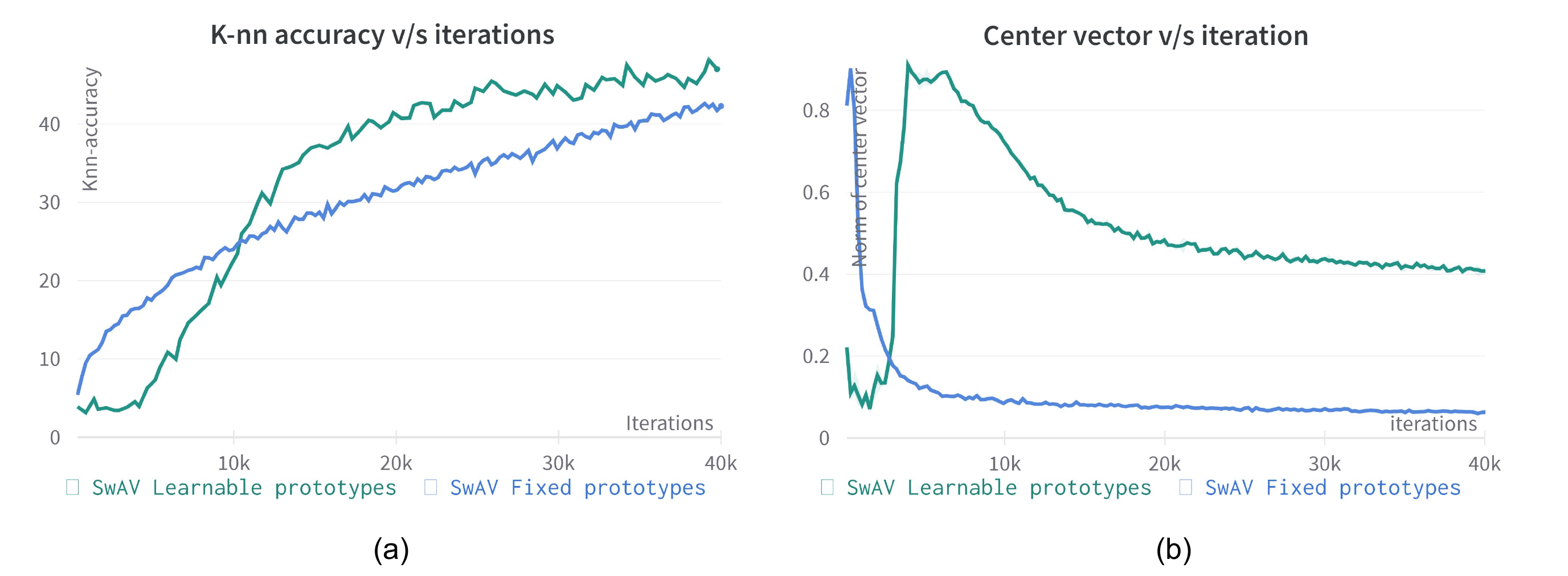}
\caption{SwAV with fixed prototype and collapse avoidance as an inductive bias.}
\label{fig:swav_fixed_acc_cv}
\end{minipage}
\quad
\begin{minipage}[b]{0.45\linewidth}
\centering
\includegraphics[width=\linewidth]{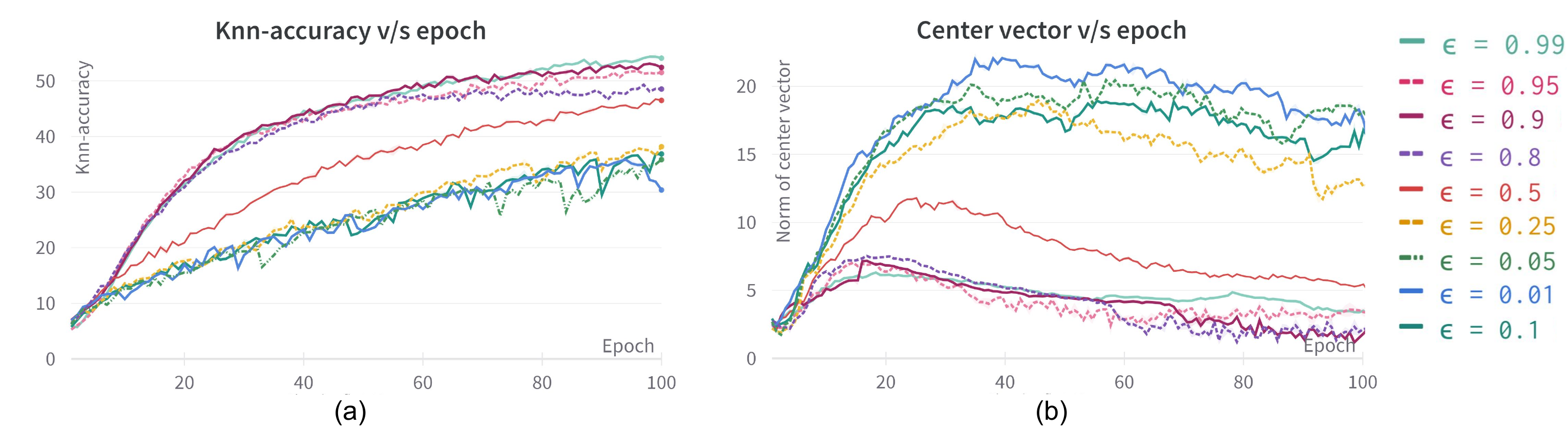}
\caption{Analyzing the relation between different values of $\epsilon$ in EMA, on the center vector of BYOL and knn-classification on Imagenet100.}
\label{fig:byol_ema_cv}
\end{minipage}
\label{fig:both}
\end{figure}

\citet{caron2020unsupervised} show that SwAV even with fixed prototypes can learn a rich feature space, resulting in a downstream performance comparable to learnable prototypes.
We analyze this fixed prototype model to investigate it in the purview of our framework. 
Figure~\ref{fig:swav_fixed_acc_cv} shows, that when the prototypes are randomly and uniformly initialized on a unit hypersphere, i.e. the $E_{z\in \text{prototypes}}[z] = 0$, the center vector magnitude is zero by design, as an inductive bias, and hence collapse is avoided. However, the manifold of the random initialized prototype space may not coincide with the natural manifold of the data semantics, hence the performance of the learnable prototypes are better than fixed.

\vspace{-0.5em}
\subsection{BYOL, is high momentum important?}
\vspace{-0.5em}
Based on our formulation, high value of momentum $\epsilon$ is important in order to confine with the initial uniform distribution of the data in the feature space, i.e. $\mathbb{E}_x[z] \approx 0$, see Section~\ref{sec:BYOL}. Here, we analyze this hypothesis, by examining the center vector and performance for different values of momentum, $\epsilon$. Figure~\ref{fig:byol_ema_cv} shows that lower momentum leads to higher center vector magnitudes, leading to instability and low knn-accuracies, while higher momentum leads to vice-versa, verifying the relation between center vector and stability thus performance in BYOL.

\section{Related work}

Non-contrastive methods like, SwAV \citep{caron2020unsupervised}, BYOL \citep{grill2020bootstrap}, Barlow-twins 
\citep{zbontar2021barlow}, SimSiam \citep{chen2020exploring}, and DINO \citep{caron2021emerging} eliminate the need for negative exemplars and use a Siamese-like architecture. An online stream (student-stream) learns through gradient-based optimization, while an offline stream (teacher-stream) computes parameters based on the student's stream without direct gradient involvement.

Some of the earlier work attempting to understand the lack of collapse of non-contrastive SSL includes \citet{tian2021understanding}, which explains the role of predictor in SimSiam as learning the eigen-space of  the feature vectors. \citet{zhang2022does} introduce negative gradient of center vector as the collapse avoidance technique, however they explain it only empirically and only for SimSiam. \citet{garrido2022duality} propose a dual relation between SimCLR and ViCReg \citep{bardes2021vicreg}. While these methods provide insights about the SSL working mechanisms, they are limited to specific methods or to empirical analysis. In this work we attempt to provide a unified framework that generalizes over different contrastive and non-contrastive methods. For a detailed literature review, please refer to the supplementary.

\section{Conclusion}

We propose a framework for collapse avoidance in self-supervised representation learning based on center vectors. The center vector magnitude needs to be minimized to prevent feature collapse, making self-supervised feature learning an optimization problem of maximizing invariance and minimizing the expected representation. Existing self-supervised techniques can be reformulated in terms of center vector minimization. Empirical analysis on Imagenet100 and toy dataset shows that collapsed versions have higher center vector magnitudes but worse knn-classification performance compared to standard versions. 
We revisit SwAV with fixed-prototypes and Barlow-twins without decorrelation loss, known not to collapse, and explain their mechanisms within our framework.
We propose a simplified SSL method based on our framework and our empirical evaluation supports our framework.

\bibliographystyle{iclr2024_conference}
\bibliography{references}

\begin{thebibliography}{41}
\providecommand{\natexlab}[1]{#1}
\providecommand{\url}[1]{\texttt{#1}}
\expandafter\ifx\csname urlstyle\endcsname\relax
  \providecommand{\doi}[1]{doi: #1}\else
  \providecommand{\doi}{doi: \begingroup \urlstyle{rm}\Url}\fi

\bibitem[Amrani et~al.(2022)Amrani, Karlinsky, and Bronstein]{amrani2022self}
Elad Amrani, Leonid Karlinsky, and Alex Bronstein.
\newblock Self-supervised classification network.
\newblock In \emph{Computer Vision--ECCV 2022: 17th European Conference, Tel Aviv, Israel, October 23--27, 2022, Proceedings, Part XXXI}, pp.\  116--132. Springer, 2022.

\bibitem[Asano et~al.(2019)Asano, Rupprecht, and Vedaldi]{asano2019self}
Yuki~Markus Asano, Christian Rupprecht, and Andrea Vedaldi.
\newblock Self-labelling via simultaneous clustering and representation learning.
\newblock \emph{arXiv preprint arXiv:1911.05371}, 2019.

\bibitem[Bachman et~al.(2019)Bachman, Hjelm, and Buchwalter]{bachman2019learning}
Philip Bachman, R~Devon Hjelm, and William Buchwalter.
\newblock Learning representations by maximizing mutual information across views.
\newblock \emph{Advances in neural information processing systems}, 32, 2019.

\bibitem[Bachmann et~al.(2022)Bachmann, Mizrahi, Atanov, and Zamir]{bachmann2022multimae}
Roman Bachmann, David Mizrahi, Andrei Atanov, and Amir Zamir.
\newblock Multimae: Multi-modal multi-task masked autoencoders.
\newblock In \emph{Computer Vision--ECCV 2022: 17th European Conference, Tel Aviv, Israel, October 23--27, 2022, Proceedings, Part XXXVII}, pp.\  348--367. Springer, 2022.

\bibitem[Bao et~al.(2021)Bao, Dong, and Wei]{bao2021beit}
Hangbo Bao, Li~Dong, and Furu Wei.
\newblock Beit: Bert pre-training of image transformers.
\newblock \emph{arXiv preprint arXiv:2106.08254}, 2021.

\bibitem[Bardes et~al.(2021)Bardes, Ponce, and LeCun]{bardes2021vicreg}
Adrien Bardes, Jean Ponce, and Yann LeCun.
\newblock Vicreg: Variance-invariance-covariance regularization for self-supervised learning.
\newblock \emph{arXiv preprint arXiv:2105.04906}, 2021.

\bibitem[Bojanowski \& Joulin(2017)Bojanowski and Joulin]{bojanowski2017unsupervised}
Piotr Bojanowski and Armand Joulin.
\newblock Unsupervised learning by predicting noise.
\newblock In \emph{International Conference on Machine Learning}, pp.\  517--526. PMLR, 2017.

\bibitem[Caron et~al.(2018)Caron, Bojanowski, Joulin, and Douze]{caron2018deep}
Mathilde Caron, Piotr Bojanowski, Armand Joulin, and Matthijs Douze.
\newblock Deep clustering for unsupervised learning of visual features.
\newblock In \emph{Proceedings of the European conference on computer vision (ECCV)}, pp.\  132--149, 2018.

\bibitem[Caron et~al.(2020)Caron, Misra, Mairal, Goyal, Bojanowski, and Joulin]{caron2020unsupervised}
Mathilde Caron, Ishan Misra, Julien Mairal, Priya Goyal, Piotr Bojanowski, and Armand Joulin.
\newblock Unsupervised learning of visual features by contrasting cluster assignments.
\newblock \emph{Advances in Neural Information Processing Systems}, 33:\penalty0 9912--9924, 2020.

\bibitem[Caron et~al.(2021)Caron, Touvron, Misra, J{\'e}gou, Mairal, Bojanowski, and Joulin]{caron2021emerging}
Mathilde Caron, Hugo Touvron, Ishan Misra, Herv{\'e} J{\'e}gou, Julien Mairal, Piotr Bojanowski, and Armand Joulin.
\newblock Emerging properties in self-supervised vision transformers.
\newblock In \emph{Proceedings of the IEEE/CVF International Conference on Computer Vision}, pp.\  9650--9660, 2021.

\bibitem[Chen et~al.(2020)Chen, Kornblith, Norouzi, and Hinton]{chen2020simple}
Ting Chen, Simon Kornblith, Mohammad Norouzi, and Geoffrey Hinton.
\newblock A simple framework for contrastive learning of visual representations.
\newblock In \emph{International conference on machine learning}, pp.\  1597--1607. PMLR, 2020.

\bibitem[Chen et~al.(2021)Chen, Luo, and Li]{chen2021intriguing}
Ting Chen, Calvin Luo, and Lala Li.
\newblock Intriguing properties of contrastive losses.
\newblock \emph{Advances in Neural Information Processing Systems}, 34:\penalty0 11834--11845, 2021.

\bibitem[Chen \& He(2020)Chen and He]{chen2020exploring}
Xinlei Chen and Kaiming He.
\newblock Exploring simple siamese representation learning. corr abs/2011.10566 (2020).
\newblock \emph{arXiv preprint arXiv:2011.10566}, 2020.

\bibitem[Deng et~al.(2009)Deng, Dong, Socher, Li, Li, and Fei-Fei]{deng2009imagenet}
Jia Deng, Wei Dong, Richard Socher, Li-Jia Li, Kai Li, and Li~Fei-Fei.
\newblock Imagenet: A large-scale hierarchical image database.
\newblock In \emph{2009 IEEE conference on computer vision and pattern recognition}, pp.\  248--255. IEEE, 2009.

\bibitem[Doersch et~al.(2015)Doersch, Gupta, and Efros]{doersch2015unsupervised}
Carl Doersch, Abhinav Gupta, and Alexei~A Efros.
\newblock Unsupervised visual representation learning by context prediction.
\newblock In \emph{Proceedings of the IEEE international conference on computer vision}, pp.\  1422--1430, 2015.

\bibitem[Dosovitskiy et~al.(2014)Dosovitskiy, Springenberg, Riedmiller, and Brox]{dosovitskiy2014discriminative}
Alexey Dosovitskiy, Jost~Tobias Springenberg, Martin Riedmiller, and Thomas Brox.
\newblock Discriminative unsupervised feature learning with convolutional neural networks.
\newblock \emph{Advances in neural information processing systems}, 27, 2014.

\bibitem[Ermolov et~al.(2021)Ermolov, Siarohin, Sangineto, and Sebe]{ermolov2021whitening}
Aleksandr Ermolov, Aliaksandr Siarohin, Enver Sangineto, and Nicu Sebe.
\newblock Whitening for self-supervised representation learning.
\newblock In \emph{International Conference on Machine Learning}, pp.\  3015--3024. PMLR, 2021.

\bibitem[Garrido et~al.(2022)Garrido, Chen, Bardes, Najman, and Lecun]{garrido2022duality}
Quentin Garrido, Yubei Chen, Adrien Bardes, Laurent Najman, and Yann Lecun.
\newblock On the duality between contrastive and non-contrastive self-supervised learning.
\newblock \emph{arXiv preprint arXiv:2206.02574}, 2022.

\bibitem[Gidaris et~al.(2018)Gidaris, Singh, and Komodakis]{gidaris2018unsupervised}
Spyros Gidaris, Praveer Singh, and Nikos Komodakis.
\newblock Unsupervised representation learning by predicting image rotations.
\newblock \emph{arXiv preprint arXiv:1803.07728}, 2018.

\bibitem[Goyal et~al.(2021)Goyal, Caron, Lefaudeux, Xu, Wang, Pai, Singh, Liptchinsky, Misra, Joulin, et~al.]{goyal2021self}
Priya Goyal, Mathilde Caron, Benjamin Lefaudeux, Min Xu, Pengchao Wang, Vivek Pai, Mannat Singh, Vitaliy Liptchinsky, Ishan Misra, Armand Joulin, et~al.
\newblock Self-supervised pretraining of visual features in the wild.
\newblock \emph{arXiv preprint arXiv:2103.01988}, 2021.

\bibitem[Grill et~al.(2020)Grill, Strub, Altch{\'e}, Tallec, Richemond, Buchatskaya, Doersch, Avila~Pires, Guo, Gheshlaghi~Azar, et~al.]{grill2020bootstrap}
Jean-Bastien Grill, Florian Strub, Florent Altch{\'e}, Corentin Tallec, Pierre Richemond, Elena Buchatskaya, Carl Doersch, Bernardo Avila~Pires, Zhaohan Guo, Mohammad Gheshlaghi~Azar, et~al.
\newblock Bootstrap your own latent-a new approach to self-supervised learning.
\newblock \emph{Advances in neural information processing systems}, 33:\penalty0 21271--21284, 2020.

\bibitem[He et~al.(2019)He, Fan, Wu, Xie, and Girshick]{he2019momentum}
Kaiming He, Haoqi Fan, Yuxin Wu, Saining Xie, and Ross Girshick.
\newblock Momentum contrast for unsupervised visual representation learning. arxiv e-prints.
\newblock \emph{arXiv preprint arXiv:1911.05722}, 2019.

\bibitem[He et~al.(2022)He, Chen, Xie, Li, Doll{\'a}r, and Girshick]{he2022masked}
Kaiming He, Xinlei Chen, Saining Xie, Yanghao Li, Piotr Doll{\'a}r, and Ross Girshick.
\newblock Masked autoencoders are scalable vision learners.
\newblock In \emph{Proceedings of the IEEE/CVF Conference on Computer Vision and Pattern Recognition}, pp.\  16000--16009, 2022.

\bibitem[Hjelm et~al.(2018)Hjelm, Fedorov, Lavoie-Marchildon, Grewal, Bachman, Trischler, and Bengio]{hjelm2018learning}
R~Devon Hjelm, Alex Fedorov, Samuel Lavoie-Marchildon, Karan Grewal, Phil Bachman, Adam Trischler, and Yoshua Bengio.
\newblock Learning deep representations by mutual information estimation and maximization.
\newblock \emph{arXiv preprint arXiv:1808.06670}, 2018.

\bibitem[Hoffer \& Ailon(2015)Hoffer and Ailon]{hoffer2015deep}
Elad Hoffer and Nir Ailon.
\newblock Deep metric learning using triplet network.
\newblock In \emph{Similarity-Based Pattern Recognition: Third International Workshop, SIMBAD 2015, Copenhagen, Denmark, October 12-14, 2015. Proceedings 3}, pp.\  84--92. Springer, 2015.

\bibitem[Kingma \& Welling(2013)Kingma and Welling]{kingma2013auto}
Diederik~P Kingma and Max Welling.
\newblock Auto-encoding variational bayes.
\newblock \emph{arXiv preprint arXiv:1312.6114}, 2013.

\bibitem[Le et~al.(2011)Le, Karpenko, Ngiam, and Ng]{le2011ica}
Quoc Le, Alexandre Karpenko, Jiquan Ngiam, and Andrew Ng.
\newblock Ica with reconstruction cost for efficient overcomplete feature learning.
\newblock \emph{Advances in neural information processing systems}, 24, 2011.

\bibitem[Misra \& Maaten(2020)Misra and Maaten]{misra2020self}
Ishan Misra and Laurens van~der Maaten.
\newblock Self-supervised learning of pretext-invariant representations.
\newblock In \emph{Proceedings of the IEEE/CVF Conference on Computer Vision and Pattern Recognition}, pp.\  6707--6717, 2020.

\bibitem[Noroozi \& Favaro(2016)Noroozi and Favaro]{noroozi2016unsupervised}
Mehdi Noroozi and Paolo Favaro.
\newblock Unsupervised learning of visual representations by solving jigsaw puzzles.
\newblock In \emph{Computer Vision--ECCV 2016: 14th European Conference, Amsterdam, The Netherlands, October 11-14, 2016, Proceedings, Part VI}, pp.\  69--84. Springer, 2016.

\bibitem[Oord et~al.(2018)Oord, Li, and Vinyals]{oord2018representation}
Aaron van~den Oord, Yazhe Li, and Oriol Vinyals.
\newblock Representation learning with contrastive predictive coding.
\newblock \emph{arXiv preprint arXiv:1807.03748}, 2018.

\bibitem[Richemond et~al.(2020)Richemond, Grill, Altch{\'e}, Tallec, Strub, Brock, Smith, De, Pascanu, Piot, et~al.]{richemond2020byol}
Pierre~H Richemond, Jean-Bastien Grill, Florent Altch{\'e}, Corentin Tallec, Florian Strub, Andrew Brock, Samuel Smith, Soham De, Razvan Pascanu, Bilal Piot, et~al.
\newblock Byol works even without batch statistics.
\newblock \emph{arXiv preprint arXiv:2010.10241}, 2020.

\bibitem[Tian et~al.(2021)Tian, Chen, and Ganguli]{tian2021understanding}
Yuandong Tian, Xinlei Chen, and Surya Ganguli.
\newblock Understanding self-supervised learning dynamics without contrastive pairs.
\newblock In \emph{International Conference on Machine Learning}, pp.\  10268--10278. PMLR, 2021.

\bibitem[Van~Gansbeke et~al.(2020)Van~Gansbeke, Vandenhende, Georgoulis, Proesmans, and Van~Gool]{van2020scan}
Wouter Van~Gansbeke, Simon Vandenhende, Stamatios Georgoulis, Marc Proesmans, and Luc Van~Gool.
\newblock Scan: Learning to classify images without labels.
\newblock In \emph{Computer Vision--ECCV 2020: 16th European Conference, Glasgow, UK, August 23--28, 2020, Proceedings, Part X}, pp.\  268--285. Springer, 2020.

\bibitem[Van~Gansbeke et~al.(2021)Van~Gansbeke, Vandenhende, Georgoulis, and Van~Gool]{van2021unsupervised}
Wouter Van~Gansbeke, Simon Vandenhende, Stamatios Georgoulis, and Luc Van~Gool.
\newblock Unsupervised semantic segmentation by contrasting object mask proposals.
\newblock In \emph{Proceedings of the IEEE/CVF International Conference on Computer Vision}, pp.\  10052--10062, 2021.

\bibitem[Vincent et~al.(2008)Vincent, Larochelle, Bengio, and Manzagol]{vincent2008extracting}
Pascal Vincent, Hugo Larochelle, Yoshua Bengio, and Pierre-Antoine Manzagol.
\newblock Extracting and composing robust features with denoising autoencoders.
\newblock In \emph{Proceedings of the 25th international conference on Machine learning}, pp.\  1096--1103, 2008.

\bibitem[Wei et~al.(2022)Wei, Fan, Xie, Wu, Yuille, and Feichtenhofer]{wei2022masked}
Chen Wei, Haoqi Fan, Saining Xie, Chao-Yuan Wu, Alan Yuille, and Christoph Feichtenhofer.
\newblock Masked feature prediction for self-supervised visual pre-training.
\newblock In \emph{Proceedings of the IEEE/CVF Conference on Computer Vision and Pattern Recognition}, pp.\  14668--14678, 2022.

\bibitem[Weng et~al.(2022)Weng, Huang, Zhao, Anwer, Khan, and Shahbaz~Khan]{weng2022investigation}
Xi~Weng, Lei Huang, Lei Zhao, Rao Anwer, Salman~H Khan, and Fahad Shahbaz~Khan.
\newblock An investigation into whitening loss for self-supervised learning.
\newblock \emph{Advances in Neural Information Processing Systems}, 35:\penalty0 29748--29760, 2022.

\bibitem[Xie et~al.(2022)Xie, Zhang, Cao, Lin, Bao, Yao, Dai, and Hu]{xie2022simmim}
Zhenda Xie, Zheng Zhang, Yue Cao, Yutong Lin, Jianmin Bao, Zhuliang Yao, Qi~Dai, and Han Hu.
\newblock Simmim: A simple framework for masked image modeling.
\newblock In \emph{Proceedings of the IEEE/CVF Conference on Computer Vision and Pattern Recognition}, pp.\  9653--9663, 2022.

\bibitem[Zbontar et~al.(2021)Zbontar, Jing, Misra, LeCun, and Deny]{zbontar2021barlow}
Jure Zbontar, Li~Jing, Ishan Misra, Yann LeCun, and St{\'e}phane Deny.
\newblock Barlow twins: Self-supervised learning via redundancy reduction.
\newblock In \emph{International Conference on Machine Learning}, pp.\  12310--12320. PMLR, 2021.

\bibitem[Zhang et~al.(2022)Zhang, Zhang, Zhang, Pham, Yoo, and Kweon]{zhang2022does}
Chaoning Zhang, Kang Zhang, Chenshuang Zhang, Trung~X Pham, Chang~D Yoo, and In~So Kweon.
\newblock How does simsiam avoid collapse without negative samples? a unified understanding with self-supervised contrastive learning.
\newblock \emph{arXiv preprint arXiv:2203.16262}, 2022.

\bibitem[Zhang et~al.(2016)Zhang, Isola, and Efros]{zhang2016colorful}
Richard Zhang, Phillip Isola, and Alexei~A Efros.
\newblock Colorful image colorization.
\newblock In \emph{Computer Vision--ECCV 2016: 14th European Conference, Amsterdam, The Netherlands, October 11-14, 2016, Proceedings, Part III 14}, pp.\  649--666. Springer, 2016.

\end{thebibliography}
\newpage
\appendix
\counterwithin{figure}{section}

\graphicspath{ {./images/} }
\section{Formulation (Extended): SwAV and Barlow Twins}
\subsection{SwAV: Role of Sinkhorn-Knopp equipartioning}
\label{sec:swav}

SwAV is a two-stream student-teacher architecture for negative-free SSL, similar to BYOL and SimSiam, Figure~\ref{fig:all_methods}d. Both of the network streams share the same encoder weight parameters, $\theta$, with only student stream receiving the loss gradient in each training iteration. Unlike BYOL and SimSiam, it lacks a learnable predictor layer over the student stream, rather it adds a Sinkhorn-Knopp regularizer layer on the teacher network. Let the two-streams network be $f: \mathcal{X}\rightarrow \mathbb{R}^D$, $x_a$ and $x_b$ being the inputs of student and teacher networks respectively, and the resulting output for these inputs being, $z_a=\operatorname{softmax}\left(f(x_a)/\tau\right)$ and $z_b=\operatorname{softmax}\left(f(x_b)/\tau\right)$, hence the loss can be written as:
\begin{align}
 L_{\text{SwAV}}(f) =  - \mathbb{E}_{x,\omega_a,\omega_b}\left[\left\langle \texttt{sg}(\operatorname{SK}(f(x_b))),  \log\left(z_a\right)\right\rangle \right] \\ \Rightarrow 
\arg\min_f L_{\text{SwAV}}(f) = \arg\min_f - \mathbb{E}_{x,\omega_a,\omega_b}\left[\left\langle\texttt{sg}(\operatorname{SK}(f(x_b))), f(x_a) - \tau \operatorname{LogSumExp}(f(x_a)/\tau) \right\rangle \right]
\end{align}
where  $\operatorname{SK}$ is the Sinkhorn-Knopp regularization function, with output, $z_{b_{sk}} := \operatorname{SK}(f(x_b))$.  As in the analysis of the InfoNCE loss, the limit of the $\operatorname{LogSumExp}$ term as $\tau$ approaches zero from above will be the maximum element of $f(x_a)$. This $\operatorname{SK}$ regularization, equipartitions the batch among a set of prototypes uniformly and randomly initialized on a unit hypersphere. The loss gradient of the output $f(x_a)$ being:
\begin{align}
\frac{\partial}{\partial f(x_a)}L_{\text{SwAV}}(f) = - \mathbb{E}_{\omega_b} \left[ z_{b_{\text{sk}}}  - %
e_{\max f(x_a)} \right]
\end{align}
where %
$e_{\max f(x_a)}$ is a canonical basis vector with $1$ in the index corresponding to the maximum value of $f(x_a)$. 
\begin{align}
\Rightarrow\frac{\partial}{\partial f(x_a)}L_{\text{SwAV}}(f) = -\mathbb{E}_{\omega_b}\left[s_{b_{\text{sk}}} + r_{b_{\text{sk}}}\right]  + e_{\max f(x_a)}
\end{align}

Here, $s_{b_{\text{sk}}}$ and $r_{b_{\text{sk}}}$ are the center vector and residual component of $z_{b_{\text{sk}}}$
 respectively. Since equipartitioning of $\operatorname{SK}$ function spreads the features in batch $z_B$ equally to the prototypes (which are themselves uniformly initialized over the unit hypersphere), this helps SwAV avoid collapse.  The final term in the gradient, when we take an expectation w.r.t.\ $x$ and $\omega_a$, will serve as an additional entropy regularization, which further prevents collapse. %

\subsection{Barlow twins}
\label{sec:barlow_twins}

Unlike other SSL methods, the method of Barlow twins does not just maximize the similarity between the two views of the same sample, it also imposes an orthogonality constraint between the feature dimensions of the representation space.

Given the Barlow-Twins framework, as shown in Figure~\ref{fig:all_methods}f, with $i$ and $j$ as index of $z \in \mathbb{R}^D$, the loss over a minibatch can be written as \citep{zbontar2021barlow}
\begin{align}
    L_{BT}(f) = \sum_i(1-C_{ii})^2 + \lambda \sum_i\sum_{j\neq i}C_{ij}^2\\
     = \sum_i \left( 1-\frac{\sum_m z_A^{i,m}  z_B^{i,m}}{\sqrt{\sum_m{z_A^{i,m}}} \sqrt{\sum_m{z_B^{i,m}}}}\right)^2 + \lambda
    \left(\sum_i\sum_{j \neq i}\frac{\sum_m{z_A^{i,m}z_B^{j,m}}}{\sqrt{\sum_m{z_A^{i,m}}}\sqrt{\sum_m{z_B^{j,m}}}}\right)^2
\end{align}
Denoting $\bar{z}_a := (z_a - \mathbb{E}_{x,\omega}[z_\omega]) / \sqrt{\mathbb{E}_{x,\omega}[(z_\omega-\mathbb{E}_{x,\omega}[z_\omega])^2]}$, where division, exponentiation, and square root are taken element-wise, and taking expectations in place of a sum over a minibatch
\begin{align}
    L_{BT}(f) = \left\| 1-\mathbb{E}_{x,\omega_a,\omega_b}[\bar{z}_a \odot \bar{z}_b]\right\|_2^2 + \lambda
    \left(\left\|\mathbb{E}_{x,\omega_a,\omega_b}[\bar{z}_a \bar{z}_b^T]\right\|_{\operatorname{Fro}}^2 - \|\mathbb{E}_{x,\omega_a,\omega_b}[\bar{z}_a \odot \bar{z}_b]\|_2^2 \right) \\
    = D - \underbrace{2\mathbb{E}_{x,\omega_a,\omega_b}[\langle \bar{z}_a,\bar{z}_b\rangle] + (1-\lambda)\left\| \mathbb{E}_{x,\omega_a,\omega_b}[\bar{z}_a \odot \bar{z}_b]\right\|_2^2}_{\text{invariance terms}} + \lambda
    \underbrace{\left\|\mathbb{E}_{x,\omega_a,\omega_b}[\bar{z}_a \bar{z}_b^T]\right\|_{\operatorname{Fro}}^2}_{\text{stability term}} 
\end{align}
If we let $\lambda = \mathcal{O}(n^{-1/2})$, then the off diagonal terms of the stability term will have a similar magnitude effect as the diagonal terms.  Using Lemmas 3.1 and 3.2 of \citet{le2011ica}, the stability term can be viewed as an expectation over terms with non-matching $x$ samples, and the combination with the invariance terms approximates contrastive learning as described in Section~\ref{sec:ConstrastiveSSL}.

\renewcommand{\theequation}{S.\arabic{equation}}

\section{Experiments}

In this section, we provide further experiments to understand the relationship between collapse and center vector. In Section~\ref{sec:empirical_analysis_collapse}, we perform an empirical analysis to understand collapse, and study what affects the magnitude and direction of center vector. In Section~\ref{sec:dino}, we study DINO to analyze collapse in the absence of centering operation; in Section~\ref{sec:simsiam_natural} we extend our analysis of SimSiam collapse done on toy dataset in the main text Section~\ref{sec:simsiam_toy}, to natural image dataset; in Section~\ref{sec:simsiam_natural}, we study the importance of predictor's learning rate in SimSiam for collapse avoidance. Finally, in Section~\ref{sec:simsiam_meta}, we provide a possible explanation for the relationship between batch size and performance on downstream task for different SSL methods.

\subsection{Empirical analysis of collapse}
\label{sec:empirical_analysis_collapse}
Here we study the possible factors that can affect the magnitude of center vector and thereby the collapse of the feature embedding. Center vector is the expected representation over the augmented input data distribution, $s := \mathbb{E}_{x\sim P_x}\left[\mathbb{E}_{\omega\sim  P_\omega}\left[g(\omega(x))\right]\right]$, hence there are three possible factors affecting this expectation: a) uneven sampling of inputs distribution: this will result in a high center vector magnitude due the expectation over a smaller region of the hypersphere.
b) Non-uniform/biased initialization of the network, $f$, that will project the input samples unevenly on the unit hypersphere, thereby resulting in the high magnitude center vector corresponding to the higher population region on the unit hypersphere; and c) Augmentation being not centered across the original samples. 

To study these phenomenon, we sample $100$ $3$-dimensional(D) points, $x in \mathcal{R}^3$ from a normal distribution centered around $0$. We projected them to a $2$D unit hypersphere (a circle of radius 1), $ z \in \mathcal{S}^2$. We use a network consisting of two-linear layers, $f: x \rightarrow z$ as the projector. We use ten augmentation each for these $100$ samples. We compute an invariance loss between the augmented versions of the input samples, as discussed in Section~\ref{sec:cv_section}. We use two augmentation strategies, one where $\mathbb{E}_\omega[x_\omega]= x$, i.e. augmentation is centered around input samples in $3$D, and $\mathbb{E}_\omega[x_\omega]= x + \Vec{k}$, where the augmentations are biased towards a fixed direction. We also use two optimization methods, a mini\-batch SGD with a batch size of $50$, and batch SGD with batch size $=$ dataset size (i.e. $1000$ including augmentation). Hence, we analyze, in total, four experiments. 

\begin{figure}[h]
    \centering
    {\includegraphics[width=0.5\textwidth]{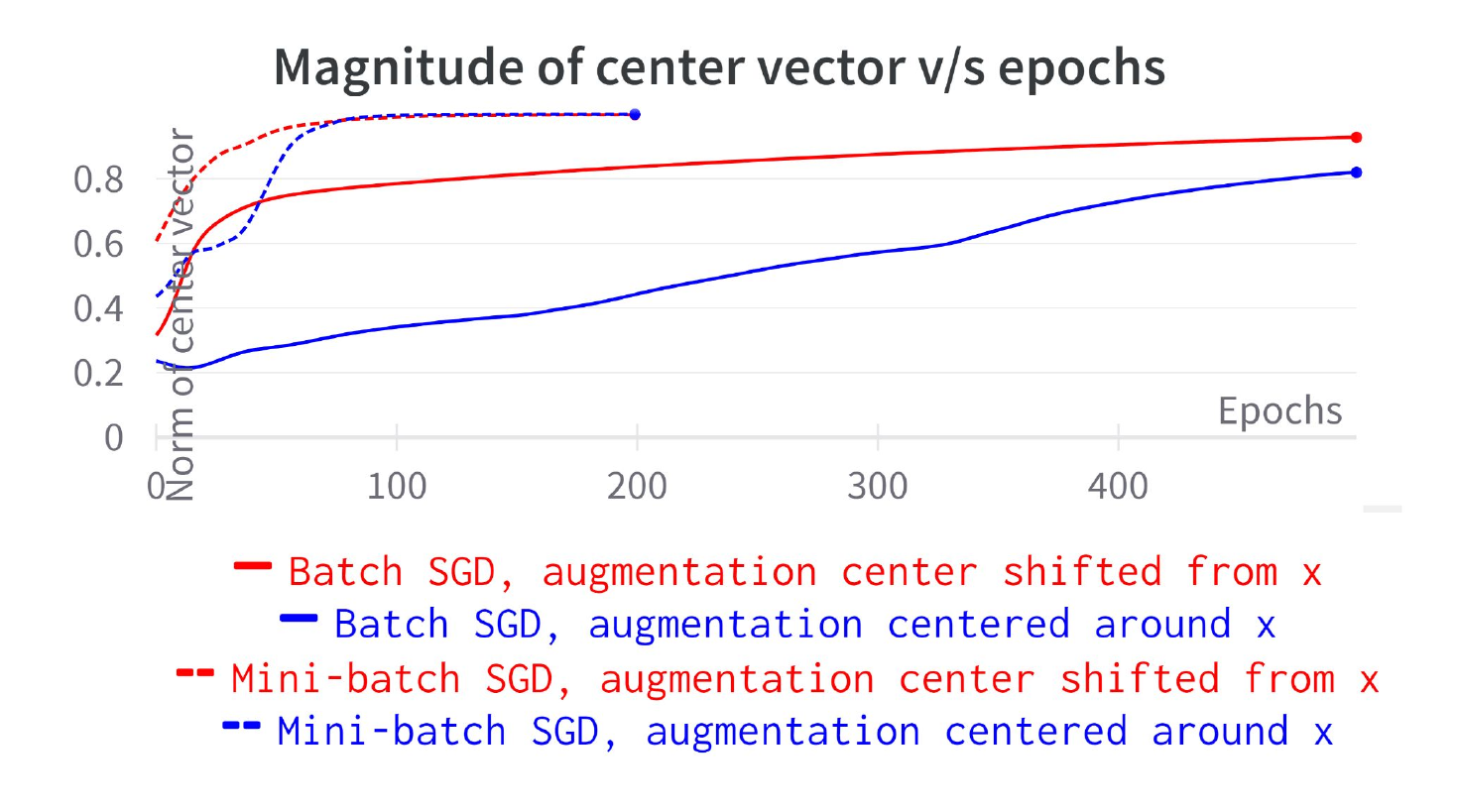}}
    \caption{Analyzing center vector magnitude for mini-batch and batch SGD optimization of invariance loss, for augmentation centered around x, and for augmentation centered shifted from x.}
    \label{sfig:factors_affecting_cv}
\end{figure}

As only invariance loss is optimized, all four methods collapse, as shown in Figure~\ref{sfig:factors_affecting_cv}.
We can observe that when mini-batch SGD is used, the collapse is faster, than when batch SGD is used, due to uneven distribution of samples in each batch. This results in a higher center vector magnitude in the initial iteration biasing the features in a certain direction,thereby resulting higher magnitude of center vector in comparison to the case of batch SGD where the projected features are more evenly distributed along the unit circle. This results in delyed collapse. 

Secondly, we observe that even for the batch SGD the representation collapse, this can be explain due to factor b) above. The non-uniform projection due to $f$ can not guarantee $\mathbb{E}_\omega[z_\omega] = z$  and $\mathbb{E}_z[z] = 0$ even if $\mathbb{E}_\omega[x_\omega] = x$ and $\mathbb{E}_x[x] = 0$. The creates a non\-uniform feature projection for even a uniform input distribution, thereby resulting in a non-zero center vector.

\begin{figure}[h]
    \centering
    {\includegraphics[width=\textwidth]{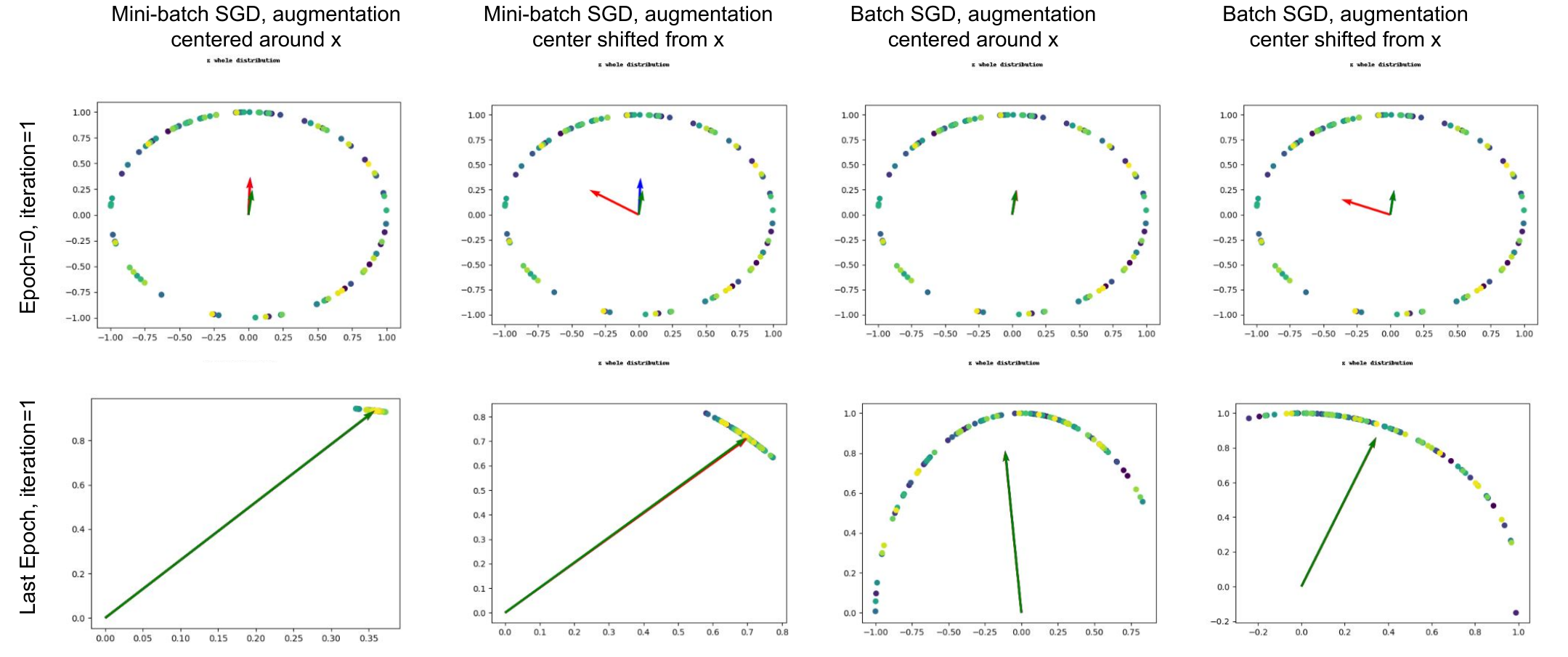}}
    \caption{\textbf{Plot of feature space $z$ for mini-batch and batch SGDs for augmentation centered around x and shifted from x:}  The \textcolor{green}{green arrow} shows the expected representation of all points in the dataset, or the center vector. The \textcolor{blue}{blue arrow} shows the expected representation $\mathbb{E}_{x\in\text{mini-batch}}[z]$, or the estimated center vector based on features $z$ of the un-augmented points. For batch SGD, the batch size is same as the population size (i.e.1000), for mini-batch SGD, the batch size is $50$. The \textcolor{red}{red arrow} shows the center vector for by taking the expectation of the features $z_\omega$ for the augmented versions of $x$. The length of these vector represent the magnitude of the center vector, and their direction shows the corresponding direction of the center vector. We can observe that for cased where the augmentation are not centered around x, the direction of the center vector is affected. We also observe that the magnitude of center vector is large for the mini-batch than the batch SGD variants. Note: i) Same seed has been used to generate the input distribution and the weight initialization of the projector function $f$. ii) Overlapping arrows may obscure the arrow color(s) underneath. iii) Last epoch for mini-batch SGD is 200 and that of batch SGD is 500, as later is more stable and requires more epochs to collapse.}
    \label{sfig:z_distribution_cv_first_epoch_last_epoch}
\end{figure}

Finally, when $\mathbb{E}_\omega[x_\omega] = x + \Vec{k}$, non-centered augmentation, the center vector angle is affected. This is due to the bias in the augmentation direction of the population, as shown in Figure \ref{sfig:z_distribution_cv_first_epoch_last_epoch}.

\begin{figure}[h]
    \centering
    {\includegraphics[width=\textwidth]{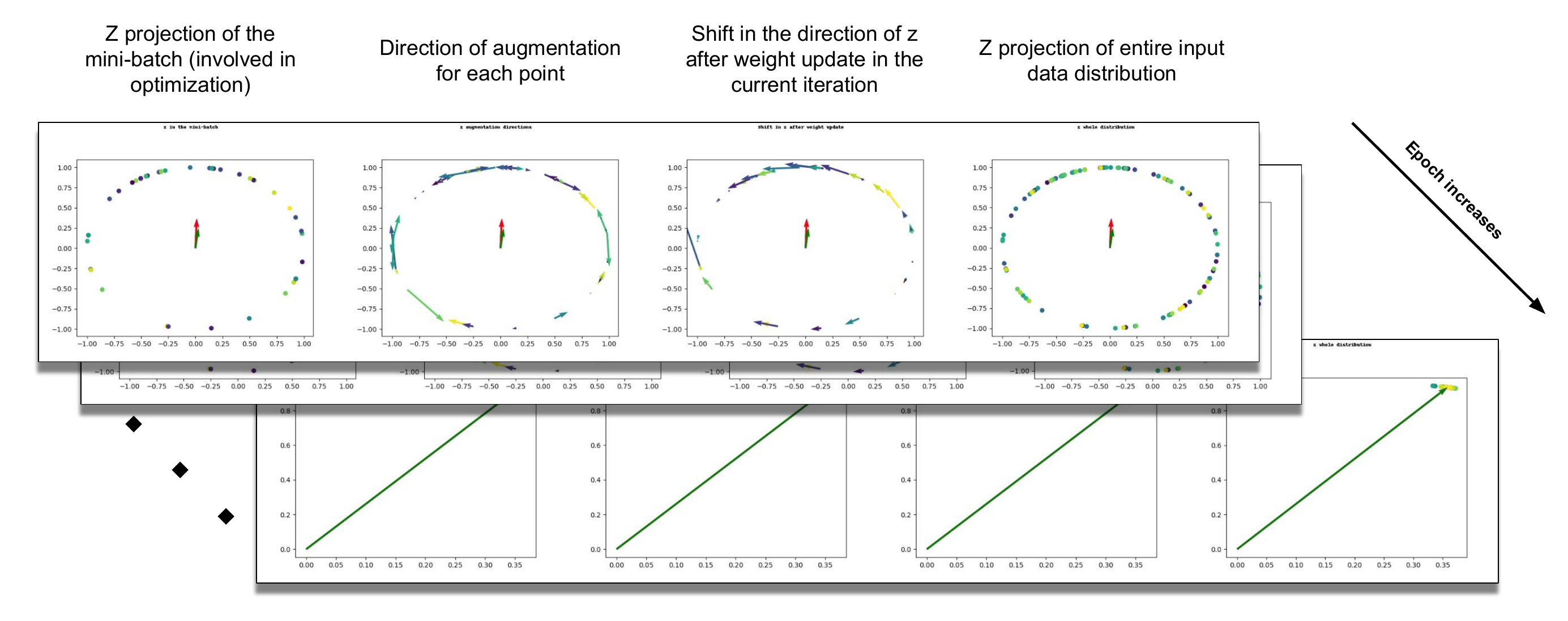}}
    \caption{\textbf{Format of the video used to demonstrate the feature space and center vectors in the four experiment discussed in Section \ref{sec:empirical_analysis_collapse} (videos\/gifs part of this supplementary file):} There are four different videos\/gifs that are submitted as a part of this supplementary; a) \textit{mini\_batch\_sgd\_centered\_augmentation};b) \textit{mini\_batch\_sgd\_shifted\_augmentation};c) \textit{batch\_sgd\_centered\_augmentation}; d)    \textit{batch\_sgd\_shifted\_augmentation}. Same seed for weight parameter of the projector, $f$, and dataset generation has been used for all four runs. Arrow definition are same as in the Figure \ref{sfig:z_distribution_cv_first_epoch_last_epoch}}
    \label{sfig:video_format}
\end{figure}

\textbf{Note:} We also, provide a video `\textit{gif}' in the supplementary that shows collapse for each of these methods as the epoch increases during the training. The format of the video is shown in Figure \ref{sfig:video_format}.

\subsection{DINO: Centering is critical}
\label{sec:dino}
\begin{figure}[h]
    \centering
    {\includegraphics[width=\textwidth]{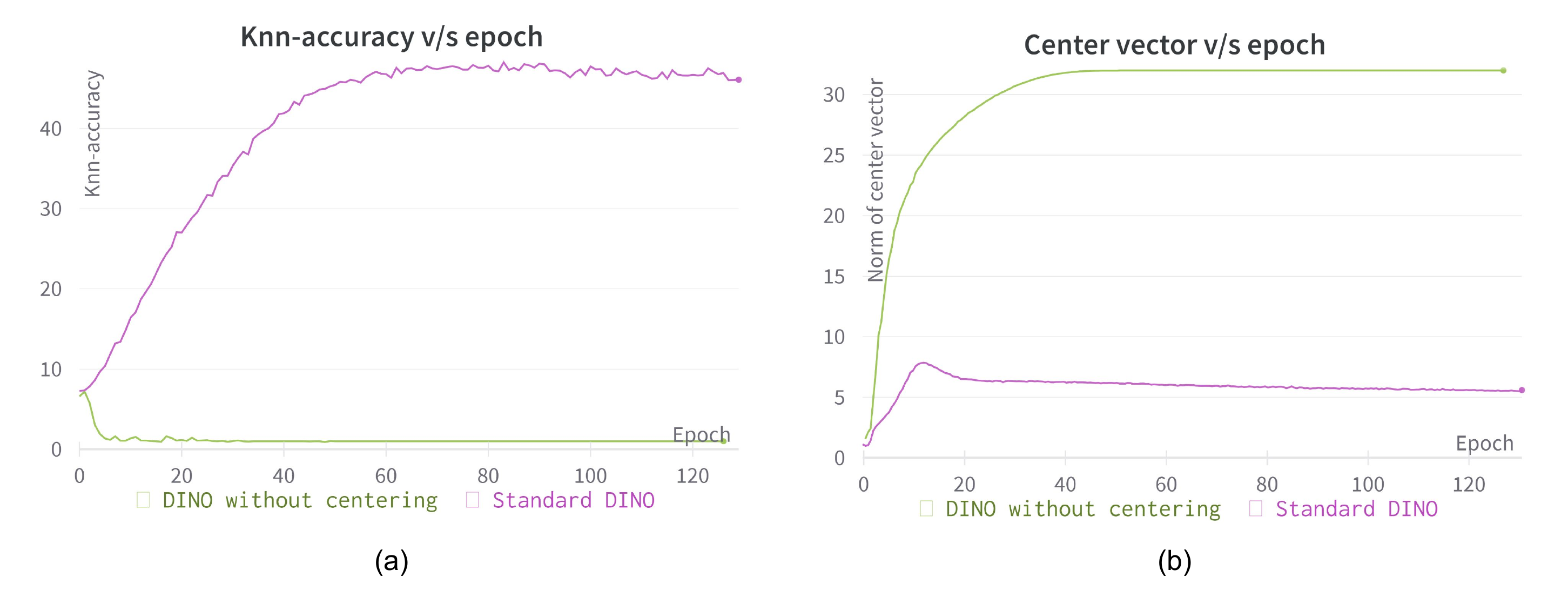}}
    \caption{Analyzing the collapse of DINO in the absence of centering, for Imagenet100: (a) shows K nearest neighbor accuracy on Imagenet100, (b) shows center vector magnitudes for the two variants. Note the rapid increase of the center vector for the collapse case, i.e. DINO without centering.}
    \label{sfig:dino_without_centering}
\end{figure}

DINO performs a centering operation on the teacher stream. In Section~\ref{sec:DINO}, we argue that since the center $C$ in DINO is updated as the EMA of the batch centers in each iteration and the teacher network is itself updated as EMA of the student network, the EMA center $C$ is approximately equivalent to the center vector, i.e. $C \approx \mathbb{E}_x[z]$. Hence subtracting the mean (EMA center) from the teacher representation $z_{b_{ema}}$ results in a small center vector and hence it avoids collapse.
To test this hypothesis, we examine the center vectors of standard DINO \citep{caron2021emerging} and DINO without centering. From the results shown in Figure~\ref{sfig:dino_without_centering}, we observe that DINO without centering experiences collapses, as indicated by the rapid increase in the magnitude of the center vector, which remains significantly larger compared to that of standard DINO. Additionally, we observe a clear inverse relationship between the magnitude of the center vector and the performance on downstream classification tasks. These findings provide empirical evidence supporting the significance of centering in DINO's training process.

\subsection{SimSiam collapses without predictor, on Natural dataset}
\label{sec:simsiam_natural}
In Section~\ref{sec:simsiam_toy} of the main text, we examined the issue of collapse in SimSiam using a toy dataset. Now, we extend our investigation to explore collapse in SimSiam with a more realistic distribution of natural images by utilizing the Imagenet100 dataset \citep{deng2009imagenet}.

\begin{figure}[h]
    \centering
    {\includegraphics[width=\textwidth]{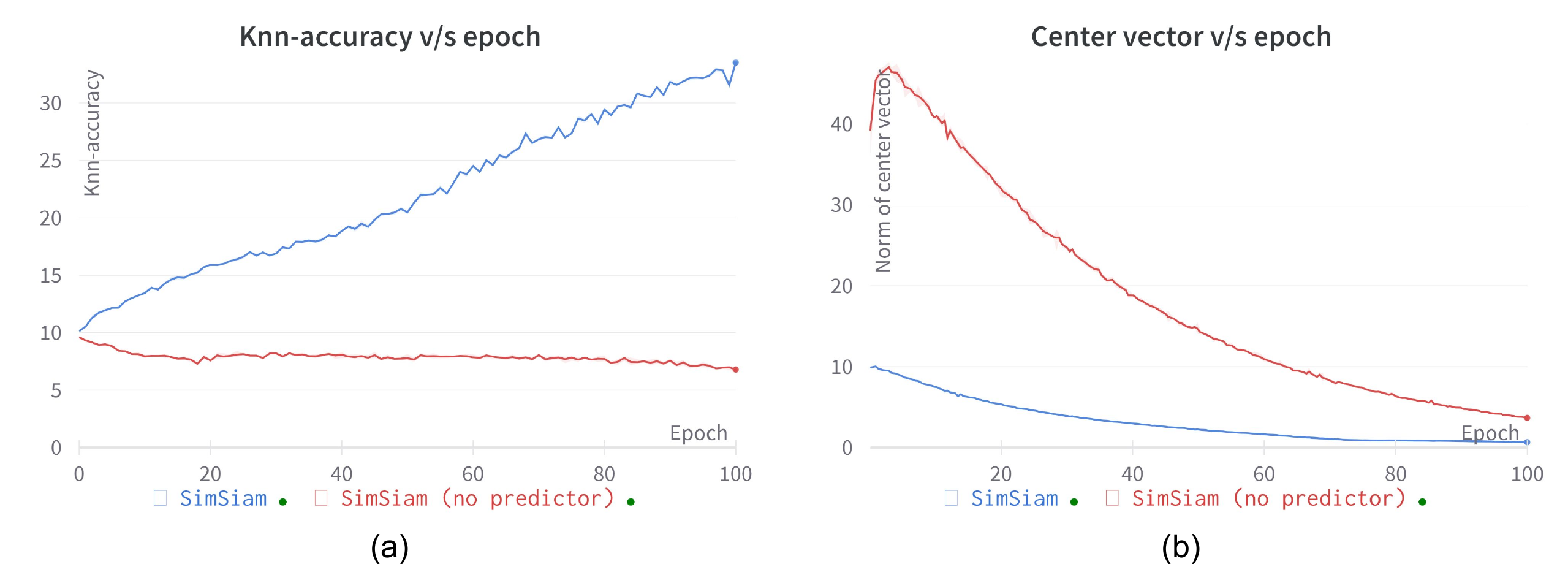}}
    \caption{Analyzing the collapse of SimSiam in the absence of predictor, for Imagenet100.}
    \label{sfig:simsiam_no_pred}
\end{figure}

For SimSiam, the predictor is critical in avoiding collapse. Based on our framework, we argue the predictor learns to map the data point to the representation space in the previous iteration, and hence minimizes the distance between the teacher and the student stream, which is equivalent to minimizing the distribution shift across the iterations. Similar to the momentum encoder in BYOL, SimSiam leverages the initial uniform distribution of the features, $\mathbb{E}_x[z] \approx 0$, to eliminate the center vector component from  the loss formulation. To study this, we analyze the center vector magnitude for the original SimSiam implementation and for SimSiam without a predictor. Figure~\ref{sfig:simsiam_no_pred} shows that without a predictor, SimSiam collapses, as the center vector magnitude is significantly larger than that for the SimSiam original implementation. This empirically shows the relation between SimSiam stability and the center vector magnitude.

\subsection{SimSiam: Importance of high learning rate of predictor}
\label{sec:simsiam_lr}

In Section~\ref{sec:simsiam}, we investigated the stability of SimSiam and derived that its loss is causally dependent on the previous iteration, as defined by equations 24 and 25. This derivation stems from the understanding that SimSiam necessitates a high learning rate for the predictor to swiftly adapt to the encoder's representation, as discussed by \citet{chen2020exploring}. Equation 18 captures this concept by stating that the predictor at iteration t projects $z_a^{t-1}$ to $z_b^{t-1}$.

To empirically validate this premise, we analyze the downstream performance of SimSiam under different predictor learning rates.

\begin{figure}[h]
    \centering
    {\includegraphics[width=0.8\textwidth]{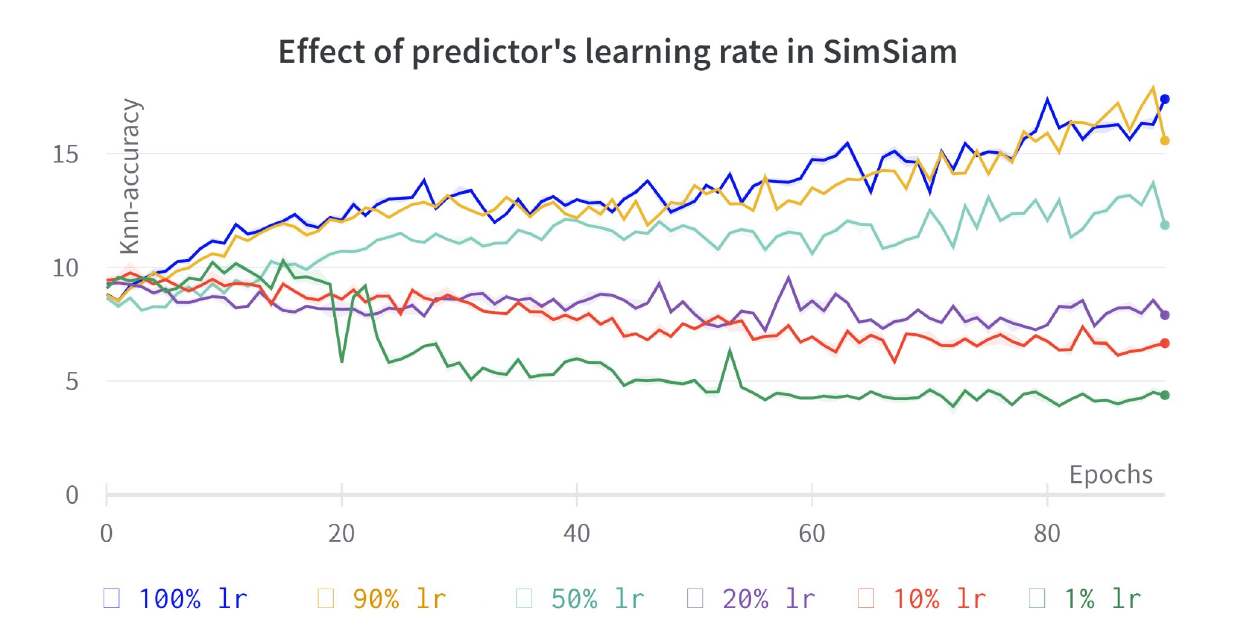}}
    \caption{Analyzing the effect of predictor's learning rate on downstream classification accuracy in SimSiam: $\%$ lr denotes the percentage of base lr for standard SimSiam being used for the experiment. We change the learning rate from $1\%$ to $100\%$. As it can be seen, for lower learning rates the model collapses, and for higher learning rates the model's downstream classification accuracy improves. Note: $100\%$ lr denotes the standard SimSiam model, and $10\%$ lr, should be similar to the predictor's learning rate in BYOL, as discussed in \citet{grill2020bootstrap}, and \citet{chen2020exploring} }.
    \label{sfig:simsiam_pred_lr_acc}
\end{figure}

We modify the standard SimSiam and multiply the predictor learning rate by $x\%$ (where $x \leq 100$), with $100\%$ meaning standard SimSiam.  We observe a collapse in SimSiam when the predictor learning rate is set to $1\%$. Conversely, we notice an improvement in downstream classification performance as the learning rate increases, as shown in Figure~\ref{sfig:simsiam_pred_lr_acc}

It is worth noting that BYOL \citep{grill2020bootstrap} employs a predictor learning rate that is 10 times smaller than that of SimSiam, which potentially explains the need for an additional exponential moving average (EMA) teacher to avoid collapse in BYOL.

\subsection{Robustness of SimSiam to batch-sizes: A meta study}
\label{sec:simsiam_meta}
In the existing literature, we observe that for SimCLR \citep{chen2020simple}, BYOL \citep{grill2020bootstrap}, DINO \citep{caron2021emerging}, Barlow-twins \citep{zbontar2021barlow}, and SwAV \citep{caron2020unsupervised}, the downstream performance is more affected by the pretraining batch-size, in comparison to SimSiam \citep{chen2020exploring}. This can be explained as SimCLR requires a large batch size to sample hard negatives, which is required to cancel out the positive center vector, Section~\ref{sec:ConstrastiveSSL}. For smaller batches in BYOL and DINO, the EMA is more affected by the batch centers computed over the small batches that are non-representative of the global distribution of the samples in the dataset. Smaller batch sizes also mean more updates of EMA, and more noising of the initial uniform distribution with zero center vector.
For SwAV, a larger batch size leads to better equipartitioning in Sinkhorn-Knopp regularization, with a larger number of prototypes getting involved in the loss computation and resulting update of prototype parameters.
Finally, in our study, we see the relation between off-diagonal elements of the cross-correlation matrix in Barlow twins and the negative pairs in contrastive SSL. Hence, a larger batch size, implicitly provides a better estimation of the center vector through the dual formulation of Barlow-twins as contrastive SSL, see Section~\ref{sec:barlow_twins}; also discussed in \citet{garrido2022duality}.

SimSiam is the only method in this study, which derives the center vector directly through a learnable layer, i.e. the prototype layer, and doesn't depend on the explicit batch samples for center vector computation, see Section~\ref{sec:simsiam}. Hence, SimSiam is more robust for different batch sizes.

\section{Related work (Extended)}

\subsection{Different self-supervised methods}

In this work we investigate the stability mechanism of self-supervised learning (SSL) methods, hence it is imperative to look at some  key prior work in the domain of visual representation learning in the absence of image labels.

Early SSL approaches aimed to learn informative image features by formulating pretext tasks that can be solved without labels. Autoencoders (AEs) and variational autoencoders (VAEs) are notable examples. AEs \citep{vincent2008extracting} learn to encode the input image into a compressed latent representation, which is then used for reconstruction. VAEs \citep{kingma2013auto} further enhance this framework by introducing a Gaussian prior in the latent space, enabling the generation of diverse outputs during the reconstruction process. Other methods like Examplar CNNs \citep{dosovitskiy2014discriminative} learn representations to classify each image as an individual class, thereby learning invariance against image augmentation introduced during training. Learning to predict rotation \citep{gidaris2018unsupervised}, relative patch location \citep{doersch2015unsupervised}, solving jigsaw of image patches \citep{noroozi2016unsupervised}, colorization of gray-scale image \citep{zhang2016colorful}, have all been used as proxy objectives, called pretext tasks. While these pretext tasks enable the learning of meaningful representations, it is worth noting that the resulting representations tend to be highly correlated with the specific pretext task and may not generalize well to other downstream tasks. These methods  are resilient to feature collapse. Since these methods are trained to predict the pretext task, a collapse in the feature space would result in an increase in the loss associated with the pretext task. This incentivizes the model to maintain diverse and informative representations throughout the training process.

An improvement over task-based representations emerged by directly optimizing image features without relying on predicting pretext tasks. Contrastive SSL methods aim to minimize the distance between two views of the same image (a positive pair) in the feature space while maximizing the distance between two different images (a negative pair). One of the pioneering works in incorporating the contrastive objective for SSL is PiRL \citep{misra2020self}].

Another approach, MOCO \citep{he2019momentum}, employs memory banks as an alternative to using large batch sizes. Contrastive predictive coding (CPC) \citep{oord2018representation} introduces the use of InfoNCE as an objective, while SimCLR, \citep{chen2020simple} introduces and analyzes strong augmentation techniques for sampling different views of the data. Other methods, such as Deep InfoMax (DIM) \citep{hjelm2018learning} and AMDIM \citep{bachman2019learning}, propose a part-vs-whole objective to maximize the agreement between image patches in the two views of the data.

These methods require hard-negative  sampling and large batches, making training computationally expensive and slow. This can be solved by non-contrastive methods that do not require a negative exemplar to avoid collapse. 
SwAV \citep{caron2020unsupervised} is a non-contrastive approach that utilizes Sinkhorn-Knopp regularization to perform online clustering in the output feature space. It aims to maximize the agreement between the student stream and the teacher stream regarding the assigned clusters.
BYOL \citep{grill2020bootstrap} incorporates a predictor on top of the student stream, and its optimization function minimizes the distance between the student's predicted output and the teacher's output. The loss gradient updates the student network, while the weights of the teacher network are updated using the exponential moving average (EMA) of the student network's weights.

SimSiam \citep{chen2020exploring}, a generalized version of BYOL, eliminates the EMA teacher update and directly updates the teacher with the weights of the student. The avoidance of collapse in BYOL and SimSiam is attributed to the asymmetry in their Siamese architecture.
SimSiam argues that the predictor learns the augmentation of the input data, effectively performing an implicit expectation maximization step in each training iteration. On the other hand, DINO \citep{caron2021emerging} demonstrates that using only EMA teacher update with centering and sharpening operation can still lead to learning non-collapsed solutions, in the absence of a predictor network.
Barlow-Twins \citep{zbontar2021barlow}, another non-contrastive technique, minimizes the distance between the covariance matrix of the two Siamese streams and the identity matrix, thereby maximizing the orthogonality of the feature components. This approach also learns non-collapse solutions.

These non-contrastive techniques employ different optimization functions to learn features that result in stable solutions. Consequently, studying the underlying mechanisms that prevent collapse in these techniques poses an interesting problem. In this work, we investigate these methods using a common framework and aim to establish a relationship between them.

Finally, there are methods that directly optimize to a fixed output space, RGB \citep{bao2021beit, he2022masked, xie2022simmim}, HOG \citep{wei2022masked}, etc, called mask-image-models (MIMs). These methods are inspired by language models where the sentence semantics can be learned by optimizing to predict masked word tokens in the input sentence, using transformers. These models do not directly optimize for a loss in their feature space, rather the reconstruction loss in the output space. As such they don't suffer from embedding collapse. The output space of these models are RGB, HOG, and discrete image tokens, and hence provide a natural output distribution to learn, in comparison to the contrastive and non-contrastive models where the output feature space is not predefined. Hence our main focus in this work lies on contrastive and non-contrastive techniques.

\subsection{Analyzing the learning dynamics}
SimSiam \citep{chen2020exploring} attributes the collapse avoidance to predictor and stop-gradient, with an explanation that the predictor learns an expectation over augmentations of the input image, with stop-gradient enabling an implicit  Expectation-Maximization (EM) like algorithm. Further investigation of SimSiam by \citet{zhang2022does} challenges this claim by proposing a negative center vector gradient for collapse avoidance and providing empirical analysis to prove their hypothesis for SimSiam. 
Another interesting study \citep{tian2021understanding} for SimSiam and BYOL, suggests predictor learns the eigenspace of the output representation space, in order to avoid collapse. While these studies primarily focus on SimSiam, our work goes beyond that. We provide both empirical analysis and a general theoretical framework that applies not only to SimSiam but also to other self-supervised learning (SSL) techniques.

In a recent study by \citet{garrido2022duality}, it is demonstrated that the loss formulation of dimensional contrastive methods such as Barlow Twins is equivalent to sample contrastive methods like SimCLR, with the addition of some constant terms. This finding is based on two lemmas from Le et al. (2011), which establish the equivalence between the orthogonality cost of the correlation matrix and the gram matrix, considering whitened input data.

Inspired by this insight, we also leverage lemmas 3.1 and 3.2 discussed in Le et al. (2011) to develop a formulation for Barlow Twins. Our aim is to demonstrate that Barlow Twins follows the same principles of center vector as contrastive methods, thereby enabling collapse avoidance. By building upon these lemmas, we provide a theoretical framework that aligns Barlow Twins with the foundations of contrastive methods.

The idea of studying the output space with a uniform distribution constraint in the context of contrastive self-supervised learning (SSL) has been explored by \citet{chen2021intriguing}. In Section~\ref{sec:swav}, we investigate SwAV with fixed prototypes, similar investigation of learning representation against uniform noise comes from \citet{weng2022investigation,bojanowski2017unsupervised}, with similar essence in Exemplar CNNs \citep{dosovitskiy2014discriminative} with categorical distributions. \citep{ermolov2021whitening} provides a similar constraint of whitening the representation space as a constraint for collapse avoidance.

\subsection{Limitation:} We focused on popular SSL techniques, but there are other interesting works in the field that we did not cover \citep{bardes2021vicreg, misra2020self, he2019momentum, amrani2022self, goyal2021self, caron2018deep, asano2019self}. Our framework can potentially be applied to those methods as well. The center vector, which represents the overall dataset without individual sample semantics, may hinder the effectiveness of semantically meaningful features like residual vectors. Therefore, removing the center vector constraint should be a crucial step in learning feature representations. However, due to space and resource constraints, we could not explore other interesting methods in the literature.

\end{document}